\documentclass[english]{article}

\usepackage{geometry}
\usepackage[T1]{fontenc}
\usepackage[latin9]{inputenc}
\usepackage{bm}
\usepackage{amsmath,mathtools}
\usepackage{amssymb}
\usepackage[unicode=true,
 bookmarks=false,
 breaklinks=false,pdfborder={0 0 1},colorlinks=false]
 {hyperref}
\hypersetup{colorlinks,citecolor=blue,filecolor=blue,linkcolor=blue,urlcolor=blue}

\makeatletter
\usepackage{amsthm}
\usepackage{cite}  
\usepackage{comment}
\usepackage{natbib}
\usepackage{booktabs}
\usepackage{mathabx}
\usepackage{graphicx}

\usepackage[linesnumbered,ruled,vlined]{algorithm2e}

\SetCommentSty{mycommfont}
\usepackage{algorithmic}

\usepackage{float}
\usepackage{multirow}
\usepackage{dsfont}
\usepackage{tcolorbox}
\usepackage{color}
\usepackage{subcaption}

\newtheorem{theorem}{Theorem}
\newtheorem{corollary}{Corollary}

\title{Connective Viewpoints of Signal-to-Noise \\ Diffusion Models }

\author{
Khanh Doan\\VinAI Research
\and Long Tung Vuong\\Monash University
\and Tuan Nguyen\\Monash University
\and Anh Tuan Bui\\Monash University
\and Quyen Tran\\VinAI Research
\and Thanh-Toan Do\\Monash University
\and Dinh Phung\\Monash University
\and Trung Le\\Monash University
}

\date{}

\begin{document}

\maketitle

\begin{abstract}
Diffusion models (DM) have become fundamental components of generative models, excelling across various domains such as image creation, audio generation, and complex data interpolation. Signal-to-Noise diffusion models constitute a diverse family covering most state-of-the-art diffusion models. While there have been several attempts to study Signal-to-Noise (S2N) diffusion models from various perspectives, there remains a need for a comprehensive study connecting different viewpoints and exploring new perspectives. In this study, we offer a comprehensive perspective on noise schedulers, examining their role through the lens of the signal-to-noise ratio (SNR) and its connections to information theory. Building upon this framework, we have developed a generalized backward equation to enhance the performance of the inference process.      


\end{abstract}

\section{Introduction}
\label{2_introduction}
Diffusion models (DM) have become a fundamental part of generative models, which excel in various domains, including creating images, generating audio, and interpolating complex data. The foundational framework for these models was introduced by \cite{sohl2015deep}, and \cite{ho2020denoising} further refined it with Denoising Diffusion Probabilistic Models (DDPMs). DDPMs add noise to data iteratively and learn to reverse this process, allowing them to model data distributions effectively.

Signal-to-Noise (S2N) diffusion models \cite{Kingma2021, kingma2024understanding} constitute an extensive class of diffusion models encompassing various other models such as variance-preserving (VP) and variance-exploding (VE) DM \cite{song2020score}, iDDPM \cite{nichol2021improved}, DDPM \cite{ho2020denoising}, EDM \cite{karras2022elucidating}, and continuous variation models \cite{Kingma2021, kingma2024understanding}. Numerous efforts have been made to study Signal-to-Noise diffusion models from various perspectives. Notably, \cite{Kingma2021} began with a discrete S2N diffusion model, developed its variational-based backward inference, and finally examined the asymptotic behavior as the number of time steps approaches infinity, resulting in a continuous variational DM. Building on the development of continuous variational DM, \cite{kingma2024understanding} further investigated S2N diffusion models in the signal-to-noise space, identifying connections between diffusion objectives with different weighting formulas and simple data augmentation techniques. Additionally, \cite{kong2023informationtheoretic} developed an information-theoretic viewpoint for S2N diffusion models in the signal-to-noise space. However, this development is limited to very specific and simple S2N diffusion models in the signal-to-noise space.           

Moreover, \cite{Zhang2022a} devised a general backward SDE from a forward S2N SDE and explored the deterministic backward SDE flow to propose a fast sampling approach based on exponential integrators. Interestingly, it can be proven that DDIM \cite{song2020score} falls within the spectrum of this devised family. However, there has been no development for the stochastic case, which is well-known to enhance the diversity of generated images. Furthermore, to summarize, there is currently no unified study connecting Markovian continuous variational DM, non-Markovian continuous variational DM, backward/forward SDE, and the information-theoretic viewpoint of S2N diffusion models.

In this work, we propose connective viewpoints of S2N diffusion models. Specifically, our contribution can be summarized as follows:
\begin{itemize}
    \item We devise a forward SDE for S2N diffusion models and demonstrate its connectivity and consistency with the results developed in \cite{Kingma2021}. Moreover, through asymptotic analysis, we show that we can inversely recover the developed forward SDE from the formula presented in \cite{Kingma2021}.
    \item To enable sampling, drawing inspiration from \cite{Zhang2022a}, we devise a general backward SDE and an exact inference formula to transition from time step $t$ to $s$ where $s<t$. Furthermore, we develop a parameterized approximate inference formula for $s=t - \Delta t$. Interestingly, we observe that the inference formula presented in \cite{Kingma2021} aligns with our parameterized approximate inference formula.
    \item Specifically, drawing inspiration from the Non-Markovian forward process in \cite{song2020}, we develop a continuous variational diffusion model capable of exactly inducing the forward distributions. Furthermore, we devise the backward SDE corresponding to this Non-Markovian inference formula.
    \item Furthermore, drawing inspiration from \cite{Kingma2021, kingma2024understanding}, we map S2N diffusion models onto the signal-to-noise space. Within this framework, we develop an information-theoretic perspective for a general S2N diffusion model, which can be seen as a generalization of the approach presented in \cite{kong2023informationtheoretic}.
    \item Finally, we employ our parameterized approximate inference formula to sample images from existing pre-trained models. This demonstration illustrates that by selecting appropriate parameters, we can achieve higher performance than the inference baselines within the spectrum. 
    
\end{itemize}

\section{Related Work}
\label{3_related_work}
Diffusion models have rapidly become a cornerstone in the landscape of generative models, demonstrating exceptional capabilities across a variety of domains, including image synthesis, audio generation, and complex data interpolation. The foundational framework of diffusion probabilistic models was first introduced by \cite{sohl2015deep}, and this framework underwent significant refinement with \cite{ho2020denoising}, who developed Denoising Diffusion Probabilistic Models (DDPMs). DDPMs iteratively add noise to data and learn to reverse this process, effectively modeling the data distribution through a sophisticated generative procedure.

Building on this foundation, subsequent research has introduced various enhancements aimed at improving the efficiency and quality of sample generation. A key development in these variants is the introduction of adaptive noise control mechanisms, often termed noise scheduling. This control is crucial as it determines the reverse diffusion trajectory, directly influencing the fidelity and diversity of the generated samples. Among these innovations, the Score-Based Generative Model (SGM) introduced by \cite{song2019generative, song2020score} represents a significant advancement. SGMs utilize score-based methods, as formalized by Hyvärinen \cite{hyvarinen2005estimation}, to guide the reverse diffusion. These methods leverage gradients of the data distribution to adaptively refine the generative process, producing samples that more closely resemble the original distribution. This approach has proven particularly effective in enhancing the visual and auditory quality of the generated outputs.

Another influential perspective is the treatment of diffusion as Continuous Normalizing Flows (CNFs), proposed by \cite{lipman2022flow, tong2023improving}. This view interprets the diffusion process as a series of invertible transformations, facilitating smoother and more controlled transitions from noise back to data. This methodology is essential for maintaining the structural integrity of complex datasets and supports a more nuanced manipulation of the generative process.

Additionally, the precise control of noise levels, conceptualized through the Signal-to-Noise Ratio (S2N), has been the focus of several studies \cite{karras2022elucidating, Kingma2021, kingma2024understanding, nichol2021improved, song2020}. The optimization of SNR is crucial, as it impacts the clarity and sharpness of the generated samples. 
By carefully tuning the SNR during the diffusion process, the model's ability to produce high-quality outputs can be significantly improved, thus avoiding common issues such as over-smoothing or excessive residual noise, which can degrade the performance of generative models. Furthermore, fast and efficient sampling has been studied in several works, notably \cite{song2020, Zhang2022a, Song2023a, Zhang2023b}.


\section{Theory Development}
\label{4_theory_development}

\subsection{Problem Setting}
We consider the following diffusion forward process
\[
\boldsymbol{z_{t}}=\alpha\left(t\right)\boldsymbol{x}+\sigma\left(t\right)\boldsymbol{\epsilon},
\]
where $\boldsymbol{\epsilon}\sim\mathcal{N}\left(\mathbf{0},\mathbf{I}\right)$,
$\boldsymbol{x}$ is generated from a data distribution, and $\alpha,\sigma:[0,T]\rightarrow\mathbb{R}^{+}$
are two functions representing signal and noise of the forward process
with $\alpha\left(0\right)=1$ and $\lim_{t\rightarrow T}\frac{\alpha\left(t\right)}{\sigma\left(t\right)}=0$. 

We define $\lambda\left(t\right)=\log\frac{\alpha\left(t\right)^{2}}{\sigma\left(t\right)^{2}}$
specifying the log of the signal-to-noise ratio with $\lim_{t\rightarrow T}\lambda\left(t\right)=-\infty$
or very low. The above signal-to-noise (S2N) forward process can be
rewritten
\begin{equation}
z_{t}=\alpha\left(t\right)\boldsymbol{x}+\alpha\left(t\right)\exp\left\{ -\lambda\left(t\right)/2\right\} \boldsymbol{\epsilon},\label{eq:SNR}
\end{equation}
where $\lambda\left(t\right)$ is a monotonic decreasing function
from $\lambda_{max}=\lambda\left(0\right)$ and $\lambda_{min}=\lambda\left(T\right)$.

Additionally, S2N diffusion models constitute a highly diverse family of diffusion models that have achieved state-of-the-art performance in practice, as summarized in Table \ref{table:Diffusion_Variants}. 

\begin{table}[ht]
    \centering
    \caption{Noise scheduling in various diffusion model variants.} 
	\smallskip
      \renewcommand\arraystretch{1.4}
    \small{
	\resizebox{1.0\textwidth}{!}{ 
        \centering
	\begin{tabular}{lcccc}
             \hline
             & $\alpha(t)$ & $\sigma(t)$ & $\lambda(t)$ & Parameters \\
             \hline
             VP \cite{song2020score} & $\displaystyle \frac{1}{\sqrt{e^{\frac{1}{2}\beta_{d}t^{2} + \beta_{min}t}}}$ & $\displaystyle \sqrt{1 - \frac{1}{e^{\frac{1}{2}\beta_{d}t^{2} + \beta_{min}t}}}$ & $\displaystyle \log\frac{1}{e^{\frac{1}{2}\beta_{d}t^{2} + \beta_{min}t} - 1}$ &  \begin{tabular}{@{}c@{}}$\beta_{min}=0.1$ \\ $\beta_{d}=19.9$\end{tabular} \\
             \\
             VE \cite{song2020score} & 1  & $\displaystyle \sigma_{min}(\frac{\sigma_{max}}{\sigma_{min}})^{t}$  & $\displaystyle (2t-2)\log\sigma_{min} - 2t\log\sigma_{max}$  & \begin{tabular}{@{}c@{}}$\sigma_{min}=0.01$ \\ $\displaystyle \sigma_{max}=50$\end{tabular} \\
             \\
             iDDPM \cite{nichol2021improved} & $\displaystyle \frac{\cos(\frac{t+s}{1+s}\cdot\frac{\pi}{2})}{\cos(\frac{s}{1+s}\cdot\frac{\pi}{2})}$  & $\displaystyle \sqrt{1 - \frac{\cos^{2}(\frac{t+s}{1+s}\cdot\frac{\pi}{2})}{\cos^{2}(\frac{s}{1+s}\cdot\frac{\pi}{2})}}$  & $\displaystyle \log\frac{\cos^{2}(\frac{t+s}{1+s}\cdot\frac{\pi}{2})}{\cos^{2}(\frac{s}{1+s}\cdot\frac{\pi}{2}) - \cos^{2}(\frac{t+s}{1+s}\cdot\frac{\pi}{2})}$ & $s=0.008$ \\
             \\
             FM-OT \cite{lipman2022flow} & $1 - t$ & $t$ & $\displaystyle 2\log\frac{1-t}{t}$ & \\
             \hline
             
            \end{tabular}
	} 
}
	\label{table:Diffusion_Variants}
\end{table}

\subsection{The Connective Viewpoints}
\paragraph{SDE Viewpoint.}
From the definition of the forward process, we know that $q\left(\boldsymbol{z}_{t}\mid\boldsymbol{z}_{0}\right)=\mathcal{N}\left(\alpha\left(t\right)\boldsymbol{z}_{0},\sigma^{2}\left(t\right)\mathbf{I}\right)$.
To realize the general transition distribution $q\left(\boldsymbol{z}_{t}\mid\boldsymbol{z}_{s}\right)$
where $0\leq s<t\leq T$, we aim to find the SDE of the above forward process.
Let us consider the general form of SDE
\begin{equation}
d\boldsymbol{z}_{t}=f\left(t\right)\boldsymbol{z}_{t}dt+g\left(t\right)d\boldsymbol{w}_{t},\label{eq:SDE}
\end{equation}
where $\{\boldsymbol{w}_{t}:t\in[0;T]\}$ is the Brownian motion,
and $f\left(t\right),g\left(t\right)\in\mathbb{R}$. 

Denote $\Psi\left(\tau,t\right)$ as the transition function satisfying
(i) $\frac{d\Psi\left(\tau,t\right)}{dt}=-\Psi\left(\tau,t\right)f\left(t\right)\mathbf{I}$,
(ii) $\frac{d\Psi\left(\tau,t\right)}{d\tau}=\Psi\left(\tau,t\right)f\left(\tau\right)\mathbf{I}$,
and (iii) $\Psi\left(\tau,\tau\right)=\mathbf{I}$. It is obvious
that $\Psi\left(\tau,t\right)=\exp\left\{ -\int_{\tau}^{t}f\left(s\right)ds\right\} \mathbf{I}$ satisfies (i), (ii), and (iii). The distribution $q\left(\boldsymbol{z}_{t}\mid\boldsymbol{z}_{s}\right)=\mathcal{N}\left(m_{t\mid s},\Sigma_{t\mid s}\right)$
is a Gaussian distribution with $m_{t\mid s}=\Psi\left(t,s\right)z_{s}$
and $\Sigma_{t\mid s}=\int_{s}^{t}\Psi\left(t,\tau\right)^{2}g^{2}\left(\tau\right)d\tau$.
Theorem \ref{thm:sde} whose proof can be found in Appendix \ref{proof1} characterizes the SDE of the forward process of S2N diffusion models.
\begin{theorem}
\label{thm:sde}With $f\left(t\right)=\frac{d\log\alpha\left(t\right)}{dt}=\frac{\alpha'\left(t\right)}{\alpha\left(t\right)}$
and $g\left(t\right)=\sqrt{-\exp\left\{ -\lambda\left(t\right)\right\} \lambda'(t)}\alpha\left(t\right)$,
the SDE flow in (\ref{eq:SDE}) is equivalent to the S2N forward process
in (\ref{eq:SNR}). Moreover, we have the transition function $\Psi\left(\tau,t\right)=\frac{\alpha\left(\tau\right)}{\alpha\left(t\right)}\mathbf{I}$
and the transition distribution $q\left(z_{t}\mid z_{s}\right)=\mathcal{N}\left(m_{t\mid s},\Sigma_{t\mid s}\right)$
with $m_{t\mid s}=\frac{\alpha\left(t\right)}{\alpha\left(s\right)}z_{0}=\alpha_{t\mid s}z_{s}$
and $\Sigma_{t\mid s}=\alpha^{2}\left(t\right)\left[\exp\left\{ -\lambda\left(t\right)\right\} -\exp\left\{ -\lambda\left(s\right)\right\} \right]=\alpha^{2}\left(t\right)\left[\frac{1}{SNR\left(t\right)}-\frac{1}{SNR\left(s\right)}\right]$
where we define $SNR\left(t\right)=\frac{\alpha\left(t\right)^{2}}{\sigma\left(t\right)^{2}}.$ Moreover, the SDE of the forward process of S2N diffusion models has the following form
\begin{equation}
d\boldsymbol{z}_{t}=\frac{\alpha'(t)}{\alpha\left(t\right)}\boldsymbol{z}_{t}dt+\sqrt{-\exp\left\{ -\lambda\left(t\right)\right\} \lambda'(t)}\alpha\left(t\right)d\boldsymbol{w}_{t}.\label{eq:forwardSDE}
\end{equation}
\end{theorem}
\paragraph{Connection to continuous variational diffusion model.} 
The results in Theorem \ref{thm:sde} are connectable and consistent
to those in continuous variational diffusion model \cite{Kingma2021}.
It is natural to ask a question: \emph{if we have the transition probability
$q\left(z_{t}\mid z_{s}\right)=\mathcal{N}\left(m_{t\mid s},\Sigma_{t\mid s}\right)$
with $m_{t\mid s}$ and $\Sigma_{t\mid s}$ defined above, can we
get back the SDE forward in Eq. (\ref{eq:SDE})}? To this end, we
consider $q\left(z_{t+\Delta t}\mid z_{t}\right)$ and derive as follows

\begin{align*}
\boldsymbol{z}_{t+\Delta t} & =\frac{\alpha\left(t+\Delta t\right)}{\alpha\left(t\right)}\boldsymbol{\boldsymbol{z}}_{t}+\alpha\left(t\right)\sqrt{\frac{1}{SNR(t+\Delta t)}-\frac{1}{SNR(t)}}\boldsymbol{\epsilon}\\
 & =\frac{\alpha\left(t+\Delta t\right)}{\alpha\left(t\right)}\boldsymbol{z}_{t}+\frac{\alpha\left(t\right)}{\sqrt{\Delta t}}\sqrt{\exp\left\{ -\lambda\left(t+\Delta t\right)\right\} -\exp\left\{ -\lambda\left(t\right)\right\} }\left(\boldsymbol{w}_{t+\Delta t}-\boldsymbol{w}_{t}\right),
\end{align*}
thanks to $\boldsymbol{w}_{t+\Delta t}-\boldsymbol{w}_{t}=\sqrt{\Delta t}\boldsymbol{\epsilon}\sim\mathcal{N}\left(\boldsymbol{0},\Delta t\boldsymbol{I}\right)$. 

This follows that
\[
\frac{\boldsymbol{\boldsymbol{z}}_{t+\Delta t}-\boldsymbol{z}_{t}}{\Delta t}=\frac{\alpha\left(t+\Delta t\right)-\alpha\left(t\right)}{\Delta t}\frac{\boldsymbol{\boldsymbol{z}}_{t}}{\alpha\left(t\right)}+\alpha\left(t\right)\sqrt{\frac{\exp\left\{ -\lambda\left(t+\Delta t\right)\right\} -\exp\left\{ -\lambda\left(t\right)\right\} }{\Delta_t}}\frac{\boldsymbol{w}_{t+\Delta t}-\boldsymbol{w}_{t}}{\Delta t}.
\]

By taking the limit when $\Delta t\rightarrow0$, we obtain
\[
d\boldsymbol{\boldsymbol{z}}_{t}=\frac{\alpha'\left(t\right)}{\alpha\left(t\right)}\boldsymbol{\boldsymbol{z}}_{t}dt+\alpha\left(t\right)\sqrt{-\exp\left(-\lambda\left(t\right)\right)\lambda'\left(t\right)}d\boldsymbol{w}_{t}=f\left(t\right)\boldsymbol{z}_{t}dt+g\left(t\right)d\boldsymbol{w}_{t},
\]
which concurs with Eq. (\ref{eq:forwardSDE}).

\paragraph{Backward SDE.} 
In what follows, we examine the backward SDE of the forward SDE in
Eq. (\ref{eq:SDE}) \cite{Zhang2022a}, which is presented in the following theorem.
\begin{theorem}
\label{thm:backwardSDE}The backward SDE of the forward SDE in Eq.
(\ref{eq:SDE}) has the following form
\begin{align}
d\boldsymbol{\boldsymbol{z}}_{t} & =\left(f\left(t\right)\boldsymbol{z}_{t}-\frac{1+\rho^{2}}{2}g^{2}\left(t\right)\nabla_{x_{t}}\log p\left(\boldsymbol{x}_{t}\right)\right)dt+\rho g\left(t\right)d\boldsymbol{w}_{t},\label{eq:backwardSDE_s}
\end{align}
where $\rho\in\mathbb{R}$ and $\nabla_{\boldsymbol{z}_{t}}\log p\left(\boldsymbol{z}_{t}\right)$
is the score function. Moreover, if we use the score network $s_{\theta}\left(\boldsymbol{z}_{t},t\right)$
to estimate $\nabla_{\boldsymbol{z}_{t}}\log p\left(\boldsymbol{\boldsymbol{z}}_{t}\right)$
and denote $s_{\theta}\left(\boldsymbol{z}_{t},t\right)=-\sigma\left(t\right)^{-1}\epsilon_{\theta}\left(\boldsymbol{\boldsymbol{z}}_{t},t\right)=-\alpha\left(t\right)^{-1}\exp\left\{ \frac{\lambda\left(t\right)}{2}\right\} \epsilon_{\theta}\left(\boldsymbol{z}_{t},t\right)$,
the backward SDE can be rewritten as
\begin{align}
d\boldsymbol{z}_{t} & =\left(\frac{\alpha'\left(t\right)}{\alpha\left(t\right)}\boldsymbol{\boldsymbol{z}}_{t}-\frac{1+\rho^{2}}{2}\exp\left\{ \frac{-\lambda\left(t\right)}{2}\right\} \lambda'\left(t\right)\alpha\left(t\right)\epsilon_{\theta}\left(\boldsymbol{\boldsymbol{z}}_{t},t\right)\right)dt+\rho g\left(t\right)d\boldsymbol{w}_{t}.\label{eq:backwardSDE_eps}
\end{align}
\end{theorem}
We now develop the exact solution of the backward SDE in Eq. (\ref{eq:backwardSDE_eps}),
allowing us to infer or sample $\boldsymbol{z}_{s}$ from $\boldsymbol{z}_{t}$
with $s<t$ in Theorem \ref{thm:backwardSDE_exact} whose proof can be found in Appendix \ref{proof3}.
\begin{theorem}
\label{thm:backwardSDE_exact}
The exact solution of the backward SDE in Eq. (\ref{eq:backwardSDE_eps})
is

\begin{align}
    \boldsymbol{z}_{s} &=\frac{\alpha\left(s\right)}{\alpha\left(t\right)}\boldsymbol{z}_{t}-\frac{1+\rho^{2}}{2}\alpha\left(s\right)\int_{t}^{s}\exp\left\{ \frac{-\lambda\left(\tau\right)}{2}\right\} \lambda'\left(\tau\right)\epsilon_{\theta}\left(\boldsymbol{z}_{\tau},\tau\right)d\tau \nonumber\\ &+\rho\alpha\left(s\right)\int_{t}^{s}\sqrt{-\exp\left\{ -\lambda\left(\tau\right)\right\} \lambda'\left(\tau\right)}d\boldsymbol{w}.\label{eq:exact_solution}
\end{align}
\end{theorem}

Furthermore, if we consider $s=t-\Delta t$ for a small $\Delta t>0$,
we can approximate $\epsilon_{\theta}\left(\boldsymbol{z}_{\tau},\tau\right)\approx\epsilon_{\theta}\left(\boldsymbol{\boldsymbol{z}}_{t},t\right)$
for $\tau\in\left[s,t\right]$, hence leading to the following approximation
solution of the exact solution in Eq. (\ref{eq:exact_solution}) as
shown in Corollary \ref{cor:approx_sol}.
\begin{corollary}
    
\label{cor:approx_sol}If we approximate $\epsilon_{\theta}\left(\boldsymbol{\boldsymbol{z}}_{\tau},\tau\right)\approx\epsilon_{\theta}\left(\boldsymbol{z}_{t},t\right)$
for $\tau\in\left[s,t\right]$, the exact solution in Eq. (\ref{eq:exact_solution})
can be approximated as
\begin{align}
\boldsymbol{\boldsymbol{z}}_{s} & =\frac{\alpha\left(s\right)}{\alpha\left(t\right)}\boldsymbol{\boldsymbol{z}}_{t}+\frac{1+\rho^{2}}{1+\gamma}\alpha\left(s\right)\left[\exp\left\{ \frac{-1-\gamma}{2}\lambda\left(t\right)\right\} -\exp\left\{ \frac{-1-\gamma}{2}\lambda\left(s\right)\right\} \right]\exp\left\{ \frac{\gamma\lambda\left(t\right)}{2}\right\} \epsilon_{\theta}\left(\boldsymbol{\boldsymbol{z}}_{t},s\right)\nonumber \\
 & +\rho\alpha\left(t\right)\sqrt{\left[\exp\left\{ -\lambda\left(t\right)\right\} -\exp\left\{ -\lambda\left(s\right)\right\} \right]}\left(\frac{\alpha\left(s\right)}{\alpha\left(t\right)}\right)^{1-\delta}\left(\frac{\sigma\left(s\right)}{\sigma\left(t\right)}\right)^{\delta}\boldsymbol{\epsilon},\label{eq:approx_infer}
\end{align}
where $\gamma\in\mathbb{R}$, $\delta\in\mathbb{R}^{+}$, and $\boldsymbol{\epsilon}\sim\mathcal{N}\left(\boldsymbol{0},\mathbf{I}\right)$. 
\end{corollary}

We now make connection to the solution $p_{\theta}\left(\boldsymbol{\boldsymbol{z}}_{s}\mid\boldsymbol{\boldsymbol{z}}_{t}\right)$
developed in \textit{continuous variational diffusion model} \cite{Kingma2021}.
Specifically, we have $p_{\theta}\left(\boldsymbol{z}_{s}\mid\boldsymbol{z}_{t}\right)=\mathcal{N}\left(\boldsymbol{z}_{t}\mid\mu_{Q}\left(\boldsymbol{z}_{t};s,t\right),\sigma_{Q}^{2}\left(s,t\right)\mathbf{I}\right)$
where we define
\begin{align*}
\mu_{Q}\left(\boldsymbol{z}_{t};s,t\right) & =\frac{\alpha\left(s\right)}{\alpha\left(t\right)}\boldsymbol{z}_{t}+\alpha\left(s\right)\alpha\left(t\right)\left(\exp\left\{ -\lambda\left(t\right)\right\} -\exp\left\{ -\lambda\left(s\right)\right\} \right)s_{\theta}\left(\boldsymbol{z}_{s},s\right)\\
 & =\frac{\alpha\left(t\right)}{\alpha\left(s\right)}\boldsymbol{z}_{t}+\alpha\left(t\right)\left(\exp\left\{ -\lambda\left(t\right)\right\} -\exp\left\{ -\lambda\left(s\right)\right\} \right)\exp\left\{ \frac{\lambda\left(t\right)}{2}\right\} \epsilon_{\theta}\left(\boldsymbol{\boldsymbol{z}}_{s},s\right)
\end{align*}
\[
\sigma_{Q}^{2}\left(s,t\right)=\sigma_{t\mid s}^{2}\sigma_{s}^{2}/\sigma_{t}^{2}=\frac{\alpha\left(t\right)^{2}\left(\exp\left\{ -\lambda\left(t\right)\right\} -\exp\left\{ -\lambda\left(s\right)\right\} \right)\sigma\left(s\right)^{2}}{\sigma\left(t\right)^{2}},
\]
which is the variance of $q\left(\boldsymbol{z}_{s}\mid\boldsymbol{z}_{t},\boldsymbol{z}_{0}\right)$.
This further implies that
\begin{align}
z_{s} & =\mu_{Q}\left(z_{t};s,t\right)+\sigma_{Q}\left(s,t\right)\boldsymbol{\epsilon}\nonumber \\
 & =\frac{\alpha\left(s\right)}{\alpha\left(t\right)}\boldsymbol{z}_{t}+\alpha\left(s\right)\alpha\left(t\right)\left(\exp\left\{ -\lambda\left(t\right)\right\} -\exp\left\{ -\lambda\left(s\right)\right\} \right)s_{\theta}\left(\boldsymbol{z}_{s},s\right)\nonumber \\
 & +\frac{\alpha\left(t\right)\sqrt{\exp\left\{ -\lambda\left(t\right)\right\} -\exp\left\{ -\lambda\left(s\right)\right\} }\sigma\left(s\right)}{\sigma\left(t\right)}\boldsymbol{\epsilon}.\label{eq:kingma_zs}
\end{align}

It is evident that the inference formula in Eq. (\ref{eq:kingma_zs})
is a special case of our general inference formula in Eq. (\ref{eq:approx_infer})
when $\gamma=1$ and $\delta=1$. Moreover, the approximated inference
formula in Eq. (\ref{eq:approx_infer}) is only sufficiently precise
when $s=t-\Delta t$ for a small step size $\Delta t>0$. This explains
why although using $p_{\theta}\left(\boldsymbol{z}_{s}\mid\boldsymbol{z}_{t}\right)$
or the inference formula in Eq. (\ref{eq:kingma_zs}) can help us
to move from $\boldsymbol{z}_{t}$ to any $\boldsymbol{z}_{s}$ as
long as $s<t$, longer step sizes $t-s$ has more errors, hence compromising
the generation performance. 

It is appealing to answer the question: \emph{from the inference formula
in Eq. (\ref{eq:kingma_zs}), can we get back the SDE backward equation
in (\ref{eq:backwardSDE_s}) with some $\rho$}? To answer this question,
from the inference formula in Eq. (\ref{eq:kingma_zs}), we set $s=t-\Delta t$\emph{
}to gain

\begin{align*}
\frac{\boldsymbol{\boldsymbol{z}}_{t-\Delta t}-\boldsymbol{\boldsymbol{z}}_{t}}{-\Delta t} & =\frac{\alpha\left(t-\Delta t\right)-\alpha\left(t\right)}{-\Delta t\alpha\left(t\right)}\boldsymbol{z}_{t}\\
 & +\frac{\alpha\left(t-\Delta t\right)\alpha\left(t\right)\left(\exp\left\{ -\lambda\left(t\right)\right\} -\exp\left\{ -\lambda\left(t-\Delta t\right)\right\} \right)}{-\Delta t}s_{\theta}\left(\boldsymbol{\boldsymbol{z}}_{t},t\right)\\
 & +\frac{\alpha\left(t\right)}{\sigma\left(t\right)}\sqrt{\frac{\exp\left\{ -\lambda\left(t\right)\right\} -\exp\left\{ -\lambda\left(t-\Delta t\right)\right\} }{\Delta t}}\sigma\left(t-\Delta t\right)\frac{\boldsymbol{w}_{t-\Delta t}-\boldsymbol{w}_{t}}{-\Delta t}.
\end{align*}
Taking limit when $\Delta t\rightarrow0$, we gain
\begin{align*}
d\boldsymbol{z}_{t} & =\left(\frac{\alpha'\left(t\right)}{\alpha\left(t\right)}\boldsymbol{z}_{t}+\alpha^{2}\left(t\right)\exp\left\{ -\lambda\left(t\right)\right\} \lambda'\left(t\right)s_{\theta}\left(\boldsymbol{z}_{t},t\right)\right)dt+\alpha\left(t\right)\sqrt{-\exp\left\{ -\lambda\left(t\right)\right\} \lambda'\left(t\right)}d\boldsymbol{w_{t}}\\
 & =\left(f\left(t\right)\boldsymbol{z}_{t}-g\left(t\right)^{2}s_{\theta}\left(\boldsymbol{z}_{t},t\right)\right)dt+g\left(t\right)d\boldsymbol{w}_{t},
\end{align*}
which falls in the spectrum of the SDE backward equation in (\ref{eq:backwardSDE_s})
with $\rho=1$. This consolidates the consistency of the SDE and the
continuous variational approach viewpoints. 

\paragraph{Non-Markovnian Continuous Variational Model and Its SDE.}
Inspired by DDIM \cite{song2020}, we relax the Markov property in the forward process and aim to find the backward distribution $q(z_{s} | z_{t}, \boldsymbol{x})$ $(s<t)$ such that its induced marginal distribution $q(\boldsymbol{z}_s)$ coincides with the forward one. To achieve this, we consider
\[
q\left(\boldsymbol{z}_{s}\mid \boldsymbol{z}_{t},\boldsymbol{x}\right)=\mathcal{N}\left(\alpha\left(s\right)\boldsymbol{z_{0}}+\sqrt{\sigma^{2}\left(s\right)-\beta^{2}\left(s,t\right)}\frac{\boldsymbol{z}_{t}-\alpha\left(t\right)\boldsymbol{z}_{0}}{\sigma\left(t\right)},\beta^{2}\left(s,t\right)\mathbf{I}\right).
\]
We now prove that if $q\left(\boldsymbol{z}_{t}\mid\boldsymbol{x}\right)=\mathcal{N}\left(\alpha\left(t\right)\boldsymbol{x},\sigma(t)^{2}\mathbf{I}\right)$
then $q\left(\boldsymbol{z}_{s}\mid\boldsymbol{x}\right)=\mathcal{N}\left(\alpha\left(s\right)\boldsymbol{x},\sigma^{2}\left(s\right)\mathbf{I}\right)$.
Indeed, we have
\[
q\left(\boldsymbol{z}_{s}\mid\boldsymbol{x}\right)=\int q\left(\boldsymbol{z}_{s}\mid z_{t},\boldsymbol{x}\right)q\left(\boldsymbol{z}_{t}\mid\boldsymbol{x}\right)dz_{t}.
\]
From \cite{bishop2007}, the mean and variance of $\boldsymbol{z}_{s}$ can be computed as
\begin{align*}
m\left(\boldsymbol{z}_{s}\right) & =\alpha\left(s\right)\boldsymbol{x}+\sqrt{\sigma^{2}\left(s\right)-\beta^{2}\left(s,t\right)}\frac{\alpha\left(t\right)\boldsymbol{x}-\alpha\left(t\right)\boldsymbol{x}}{\sigma\left(t\right)}=\alpha\left(s\right)\boldsymbol{x}.\\
V\left(\boldsymbol{z}_{s}\right) & =\beta^{2}\left(s,t\right)\mathbf{I}+\left(\sigma^{2}\left(s\right)-\beta^{2}\left(s,t\right)\right)\frac{\sigma^{2}\left(t\right)}{\sigma^{2}\left(t\right)}\mathbf{I}=\sigma^{2}\left(s\right)\mathbf{I}.
\end{align*}
Given $q\left(\boldsymbol{z}_{T}\mid\boldsymbol{x}\right)=\mathcal{N}\left(\alpha\left(T\right)\boldsymbol{x},\sigma^{2}\left(T\right)\mathbf{I}\right)$,
we reach $q\left(\boldsymbol{z}_{t}\mid\boldsymbol{x}\right)=\mathcal{N}\left(\alpha\left(t\right)\boldsymbol{x},\sigma^{2}\left(t\right)\mathbf{I}\right),\forall t\in\left[0;T\right]$,
indicating that $q\left(\boldsymbol{z}_{s}\mid \boldsymbol{z}_{t},\boldsymbol{x}\right)=\mathcal{N}\left(\alpha\left(s\right)\boldsymbol{z}_{0}+\sqrt{\sigma^{2}\left(s\right)-\beta^{2}\left(s,t\right)}\frac{\boldsymbol{z}_{t}-\alpha\left(t\right)\boldsymbol{z}_{0}}{\sigma\left(t\right)},\beta^{2}\left(s,t\right)\mathbf{I}\right)$
is a proper backward flow. Moreover, by defining 
\[
p_{\theta}\left(\boldsymbol{z}_{s}\mid \boldsymbol{z}_{t}\right)=q\left(\boldsymbol{z}_{s}\mid \boldsymbol{z}_{t},\hat{z}_{\theta}\left(\boldsymbol{z}_{t},t\right)\right),
\]
where $\hat{z}_{\theta}\left(\boldsymbol{z}_{t},t\right)=\frac{\sigma^{2}\left(t\right)s_{\theta}\left(\boldsymbol{z}_{t},t\right)+\boldsymbol{z}_{t}}{\alpha\left(t\right)}=\frac{\boldsymbol{z}_{t}-\sigma\left(t\right)\epsilon_{\theta}\left(\boldsymbol{z}_{t},t\right)}{\alpha\left(t\right)}$
is used to predict $\boldsymbol{x}$, we reach 
\begin{align}
\boldsymbol{z}_{s} & =\alpha\left(s\right)\hat{z}_{\theta}\left(z_{t},t\right)+\sqrt{\sigma^{2}\left(s\right)-\beta^{2}\left(s,t\right)}\frac{\boldsymbol{z}_{t}-\alpha\left(t\right)\hat{z}_{\theta}\left(\boldsymbol{z}_{t},t\right)}{\sigma\left(t\right)}+\beta\left(s,t\right)\boldsymbol{\epsilon}\nonumber \\
 & =-\left(\exp\left\{ \frac{\lambda\left(s\right)}{2}\right\} -\exp\left\{ \frac{\lambda\left(t\right)}{2}\right\} \sqrt{1-\frac{\beta^{2}\left(s,t\right)}{\sigma^{2}\left(s\right)}}\right)\exp\left\{ \frac{-\lambda\left(t\right)-\lambda\left(s\right)}{2}\right\} \alpha\left(s\right)\epsilon_{\theta}\left(z_{t},t\right)\nonumber \\
 & +\frac{\alpha\left(s\right)}{\alpha\left(t\right)}\left(1+\exp\left\{ \frac{\lambda\left(t\right)-\lambda\left(s\right)}{2}\right\} \sqrt{1-\frac{\beta^{2}\left(s,t\right)}{\sigma^{2}\left(s\right)}}\right)\boldsymbol{z}_{t}+\beta\left(s,t\right)\boldsymbol{\epsilon},\label{eq:non_markov_zs}
\end{align}
where $\boldsymbol{\epsilon} \sim \mathcal{N}(\boldsymbol{0}, \mathbf{I})$.

It is appealing to ask the question: \textit{what is the SDE that corresponds to the continuous variational model in Eq. (\ref{eq:non_markov_zs})?} The following theorem answers this question.
\begin{theorem}
    \label{theo:non_Markov}
    Consider $\beta\left(s,t\right)=\sqrt{b\left(s\right)-b\left(t\right)}$ for $s<t$ with a decreasing function $b$. The SDE that corresponds to the continuous variational model in Eq. (\ref{eq:non_markov_zs}) has the following form:
    \begin{align}
d\boldsymbol{z}_{t} & =\left[\frac{\alpha'\left(t\right)}{\alpha\left(t\right)}+\frac{\lambda'\left(t\right)}{2}\exp\left\{ \frac{\lambda\left(t\right)}{2}\right\} \right]\boldsymbol{\boldsymbol{z}}_{t}-\frac{\lambda'\left(t\right)}{2}\exp\left\{ -\frac{\lambda\left(t\right)}{2}\right\} \alpha\left(t\right)\epsilon_{\theta}\left(\boldsymbol{z}_{t},t\right)dt\nonumber \\
 & +\frac{1}{2}b'\left(t\right)\exp\left\{ \frac{\lambda\left(t\right)}{2}\right\} \alpha^{-1}\left(t\right)\epsilon_{\theta}\left(\boldsymbol{z}_{t},t\right)+\sqrt{-b'\left(t\right)}d\boldsymbol{w}_{t}.\label{eq:non_markov_SDE}
\end{align}
\end{theorem}

\subsection{Transforming S2N Diffusion Models to Signal-to-Noise Space and Information Theory Viewpoint}
We denote $\tilde{\alpha}\left(\lambda\left(t\right)\right)=\alpha\left(t\right)$
(i.e., $\tilde{\alpha}=\alpha\circ\lambda^{-1}$), $\tilde{\sigma}\left(\lambda\left(t\right)\right)=\sigma\left(t\right)$
(i.e., $\tilde{\sigma}=\sigma\circ\lambda^{-1}$), and $\tilde{\boldsymbol{\boldsymbol{z}}}_{\lambda\left(t\right)}=\boldsymbol{z}_{t}$
where $\lambda\left(t\right)=\log\frac{\alpha\left(t\right)}{\sigma\left(t\right)}=\log\frac{\tilde{\alpha}\left(\lambda\left(t\right)\right)^{2}}{\tilde{\sigma}\left(\lambda\left(t\right)\right)^{2}}$.
We have the following forward process in the signal-to-noise space
\[
\tilde{\boldsymbol{\boldsymbol{z}}}_{\lambda\left(t\right)}=\tilde{\alpha}\left(\lambda\left(t\right)\right)\boldsymbol{x}+\tilde{\sigma}\left(\lambda\left(t\right)\right)\boldsymbol{\epsilon}=\tilde{\alpha}\left(\lambda\left(t\right)\right)\boldsymbol{x}+\frac{\tilde{\alpha}\left(\lambda\left(t\right)\right)}{\exp\left\{ \lambda\left(t\right)/2\right\} }\boldsymbol{\epsilon}
\]
or equivalently
\begin{equation}
    \label{eq:lambda_diffusion}
    \tilde{\boldsymbol{\boldsymbol{z}}}_{\lambda}=\tilde{\alpha}\left(\lambda\right)\boldsymbol{x}+\frac{\tilde{\alpha}\left(\lambda\right)}{\exp\left\{ \lambda/2\right\} }\boldsymbol{\epsilon},
\end{equation}
where $\lambda\in\left[\lambda_{min},\lambda_{max}\right]$ with $\lambda_{min}=\lambda\left(T\right)$
and $\lambda_{max}=\lambda\left(0\right)$. 

In the following theorem, we answer the question which pair of $\left(\alpha\left(t\right),\sigma\left(t\right)\right)$
induces the same forward process in the signal-to-noise space.
\begin{theorem}
\label{thm:equivalent}
Given $\left(\alpha_{1}\left(t\right),\sigma_{1}\left(t\right)\right)$, $\sigma_{1}\circ\lambda_{1}^{-1}=\sigma_{2}\circ\lambda_{2}^{-1}$, and $\left(\alpha_{2}\left(t\right),\sigma_{2}\left(t\right)\right)$,
if $\lambda_{1}\left(0\right)=\lambda_{2}\left(0\right)$, $\lambda_{1}\left(T\right)=\lambda_{2}\left(T\right)$,
and $\alpha_{1}\circ\lambda_{1}^{-1}=\alpha_{2}\circ\lambda_{2}^{-1}$,
the forward processes corresponding to $\left(\alpha_{1}\left(t\right),\sigma_{1}\left(t\right)\right)$
and $\left(\alpha_{2}\left(t\right),\sigma_{2}\left(t\right)\right)$
induce the same forward process in the signal-to-noise space.
\end{theorem}
\paragraph{Information-Theoretic viewpoint of S2N DM in the signal-to-noise space.}
Information theoretic viewpoint was studied in \cite{kong2023informationtheoretic} for a very simple diffusion process: $\boldsymbol{\tilde{z}}_\lambda = \sqrt{\lambda}\boldsymbol{x} + \boldsymbol{\epsilon}$. In what follows, we present the information-theoretic results for the general S2N diffusion model in the signal-to-noise space (\ref{eq:lambda_diffusion}).

Given $\boldsymbol{x}$, we define the Minimum Mean Square Error (MMSE)
for recovering $\boldsymbol{x}$ in the noisy channel
\[
\text{mmse}\left(\lambda\right):=\text{min}_{\hat{\boldsymbol{x}}\left(\tilde{\boldsymbol{z}}_{\lambda},\lambda\right)}\mathbb{E}_{p\left(\tilde{\boldsymbol{z}}_{\lambda},\boldsymbol{x}\right)}\left[\Vert\boldsymbol{x}-\hat{\boldsymbol{\boldsymbol{x}}}\left(\tilde{\boldsymbol{z}}_{\lambda},\lambda\right)\Vert_{2}^{2}\right],
\]
where $\hat{\boldsymbol{x}}\left(\tilde{\boldsymbol{z}}_{\lambda},\lambda\right)$
is referred to as a denoising function. The optimal denoising function
$\hat{\boldsymbol{x}}^{*}$ corresponds to the conditional expectation,
which can be seen using variational calculus or from the fact that
the squared error is a Bregman divergence
\[
\hat{\boldsymbol{x}}^{*}\left(\tilde{\boldsymbol{z}}_{\lambda},\lambda\right)=\text{argmin}_{\hat{\boldsymbol{x}}\left(\tilde{\boldsymbol{z}}_{\lambda},\lambda\right)}\mathbb{E}_{p\left(\tilde{\boldsymbol{z}}_{\lambda},\boldsymbol{x}\right)}\left[\Vert\boldsymbol{x}-\hat{\boldsymbol{\boldsymbol{x}}}\left(\tilde{\boldsymbol{z}}_{\lambda},\lambda\right)\Vert_{2}^{2}\right]=\mathbb{E}_{\boldsymbol{x}\sim p\left(\boldsymbol{x}\mid\boldsymbol{z}_{\lambda}\right)}\left[\boldsymbol{x}\right].
\]
Moreover, the point-wise MMSE is defined as follows:
\[
\text{mmse}\left(\boldsymbol{x},\lambda\right):=\mathbb{E}_{p\left(\tilde{\boldsymbol{z}}_{\lambda},\boldsymbol{x}\right)}\left[\Vert\boldsymbol{x}-\hat{\boldsymbol{\boldsymbol{x}}}^{*}\left(\tilde{\boldsymbol{z}}_{\lambda},\lambda\right)\Vert_{2}^{2}\right].
\]
The mutual information $\mathbb{I}\left(\boldsymbol{x},\boldsymbol{\tilde{z}}_{\lambda}\right)$ can be characterized in the following theorem.
\begin{theorem}
    \label{theo:information}
    For a general S2N DM in the general signal-to-noise space, we have
    
    (i) $\frac{d}{d\lambda}D_{KL}\left(p\left(\tilde{\boldsymbol{z}}_{\lambda}\mid\boldsymbol{x}\right)\Vert p\left(\tilde{\boldsymbol{z}}_{\lambda}\right)\right)=-\frac{D\tilde{\sigma}'\left(\lambda\right)}{2\tilde{\sigma}\left(\lambda\right)}+\frac{\left[\tilde{\alpha}'\left(\lambda\right)\tilde{\sigma}\left(\lambda\right)-\tilde{\alpha}\left(\lambda\right)\tilde{\sigma}'\left(\lambda\right)\right]\tilde{\alpha}\left(\lambda\right)}{\tilde{\sigma}^{3}\left(\lambda\right)}\text{mmse}\left(\boldsymbol{x},\lambda\right)$ where $D$ is the dimension of $\tilde{\boldsymbol{z}}_\lambda$, $D_{KL}$ is the Kullback-Leibler divergence, and $\tilde{\sigma}\left(\lambda\right)=\tilde{\alpha}\left(\lambda\right)\exp\left\{ -\lambda/2\right\} $.

    (ii) $\frac{d}{d\lambda}\mathbb{I}\left(\boldsymbol{x},\boldsymbol{\tilde{z}}_{\lambda}\right)=-\frac{D\tilde{\sigma}'\left(\lambda\right)}{2\tilde{\sigma}\left(\lambda\right)}+\frac{\left[\tilde{\alpha}'\left(\lambda\right)\tilde{\sigma}\left(\lambda\right)-\tilde{\alpha}\left(\lambda\right)\tilde{\sigma}'\left(\lambda\right)\right]\tilde{\alpha}\left(\lambda\right)}{\tilde{\sigma}^{3}\left(\lambda\right)}\text{mmse}\left(\lambda\right)$.
    
\end{theorem}
It is worth noting that our results in Theorem \ref{theo:information} can lead to those in \cite{kong2023informationtheoretic} when choosing $\tilde{\alpha}\left(\lambda\right)=\sqrt{\lambda}$ and $\tilde{\sigma}\left(\lambda\right)=1$.

\section{Experiments}
\label{5_experiments}

Inspired by the theoretical results in Section \ref{4_theory_development}, we conduct experiments to test the effectiveness of hyperparameters participating in the backward process built based on our Corollary \ref{cor:approx_sol}. Our experiment settings are organized based on work and checkpoints in EDM \cite{karras2022elucidating}.

\subsection{Deterministic sampling}
\label{exp:deterministic}

Corollary \ref{cor:approx_sol} becomes deterministic when $\rho = 0$, then the sampling process is defined as follow:

\begin{align}
\boldsymbol{\boldsymbol{z}}_{s} & =\frac{\alpha\left(s\right)}{\alpha\left(t\right)}\boldsymbol{\boldsymbol{z}}_{t}-\frac{1}{1+\gamma}\alpha\left(s\right)\left[\exp\left\{ \frac{-1-\gamma}{2}\lambda\left(t\right)\right\} -\exp\left\{ \frac{-1-\gamma}{2}\lambda\left(s\right)\right\} \right]\exp\left\{ \frac{\gamma\lambda\left(t\right)}{2}\right\} \epsilon_{\theta}\left(\boldsymbol{\boldsymbol{z}}_{t},s\right)\,\label{eq:approx_deterministic_infer}
\end{align}

The traditional Euler solver method corresponds to our specific case when $\gamma = 0$. As shown in Figure \ref{fig:Varying_gamma}, with the same number of NFEs, more negative $\gamma$ makes the outcome images blurrier, while images become sharper as $\gamma$ increases. However, too large a 
 $\gamma$ value exceeds the common range of pixel values and distorts the images.

\begin{table}[ht]
\centering
\caption{Results in FID $(\downarrow)$ for Unconditional FFHQ $(64\times64)$, Unconditional AFHQv2 $(64\times64)$ and Conditional ImageNet $(64\times64)$ settings by deterministic sampling with NFE = 79 using \cite{karras2022elucidating}'s checkpoints.}
\begin{tabular}{cccccc}
\hline
                        & \multicolumn{2}{c}{Uncond. FFHQ} & \multicolumn{2}{c}{Uncond. AFHQv2} & \multirow{2}{*}{Cond. ImageNet} \\
                        \cmidrule(lr){2-5} 
                        & VP              & VE             & VP               & VE              &                                 \\
                        \cmidrule(lr){1-6} 
Euler solver            & \textbf{3.25}            & 3.43           & 2.38             & 2.58            & 2.75                            \\
Ours $(\gamma = 0.026)$ & 3.27            & \textbf{3.39}           & \textbf{2.09}             & \textbf{2.27}            & \textbf{2.71} \\
\hline
\end{tabular}
\label{tab:Deterministic_Sampling_FFHQ_AFHQV2_FID}
\end{table}

\begin{figure*}[ht]
\centering
    \includegraphics[width=1.0\textwidth]{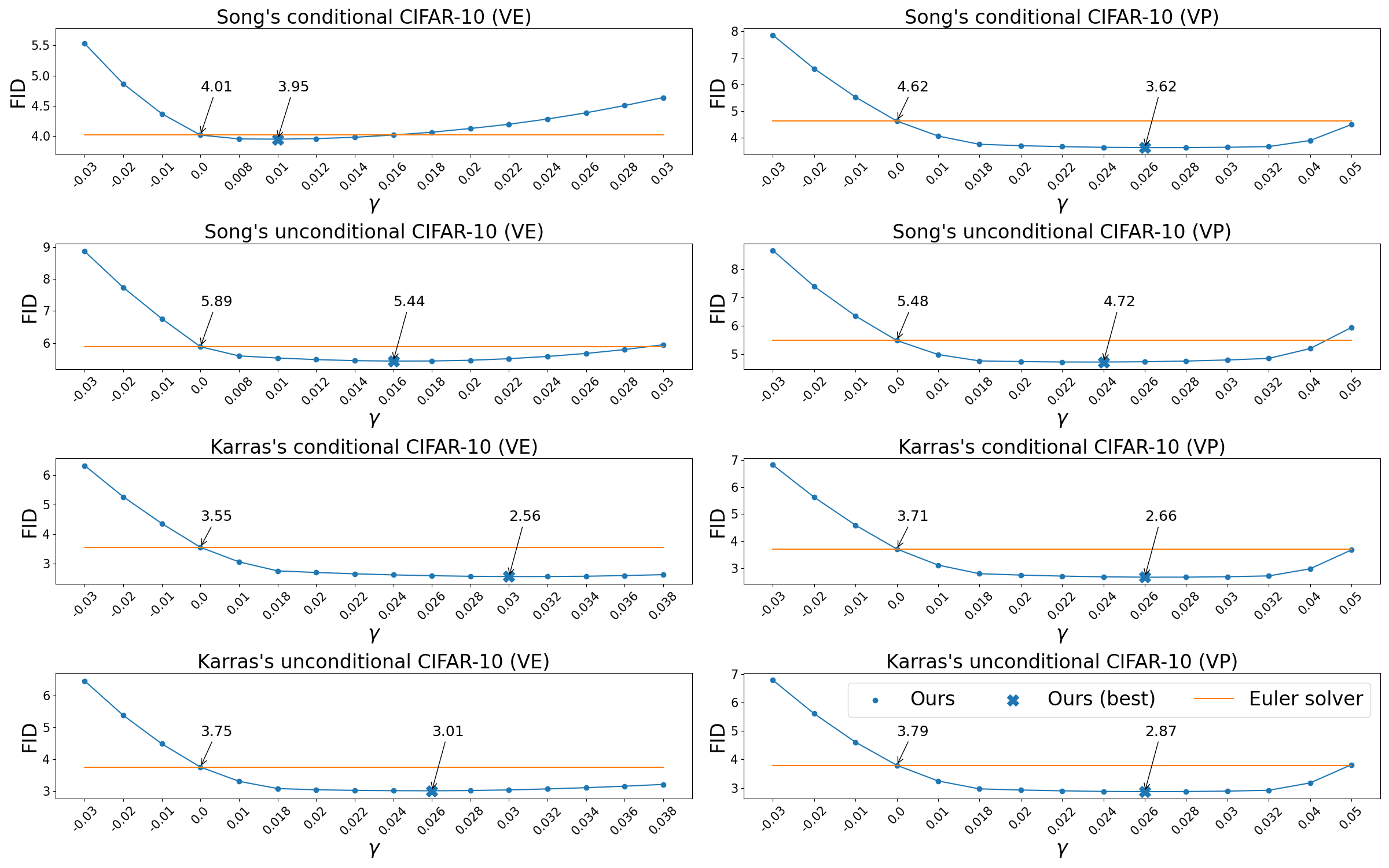}
    \caption{The FID $(\downarrow)$ of deterministic sampling at several CIFAR-10 $(32\times32)$ checkpoints when varying $\gamma$ with NFE = 35.}
    \label{fig:Deterministic_Sampling_CIFAR10_FID}
\end{figure*}

Figure \ref{fig:Deterministic_Sampling_CIFAR10_FID} represents our grid search results to find the optimal value of $\gamma$ for each CIFAR-10 $(32\times32)$ model pretrained by \cite{song2020score} and \cite{karras2022elucidating}. We observe that in all settings, the optimal $\gamma$ value, which corresponds to the best FID, is a small positive number, especially around $0.026$ for the settings used by \cite{karras2022elucidating}. Not only for CIFAR-10 $(32\times32)$, but $\gamma = 0.026$ also outperforms $\gamma = 0$ in nearly all cases for the Unconditional FFHQ $(64\times64)$, Unconditional AFHQv2 $(64\times64)$, and Conditional ImageNet $(64\times64)$, as shown in Table \ref{tab:Deterministic_Sampling_FFHQ_AFHQV2_FID}. With the FFHQ dataset, we achieve only an approximate result in VP $(3.27 > 3.25)$ and a slight improvement in VE $(3.39 < 3.43)$. Results in ImageNet also display slight improvement $(2.71 < 2.75)$. On the other hand, the evaluation for AFHQv2 shows a significant decrease in FID: $0.29$ in VP and $0.31$ in VE.

\subsection{Stochastic sampling}
\label{exp:stochastic}

    


\begin{figure}
\centering
\begin{minipage}{.432\textwidth}
  \centering
  \includegraphics[width=1.\linewidth]{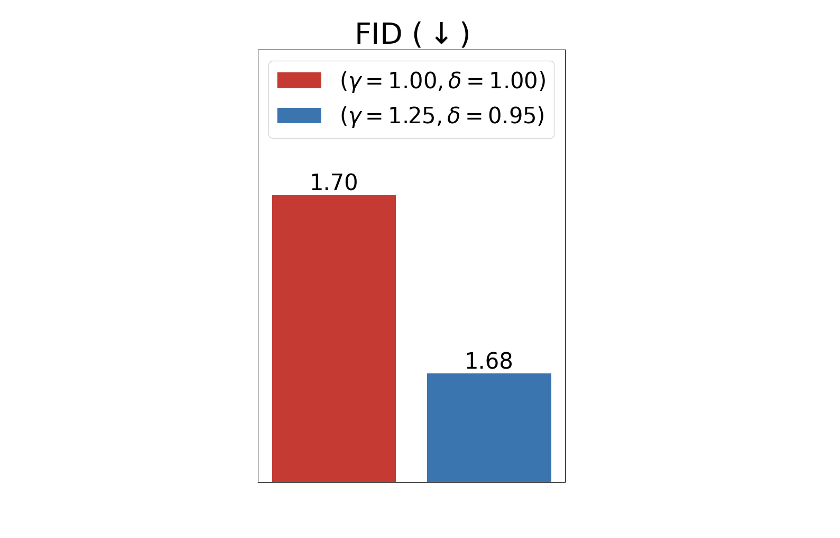}
  \captionof{figure}{FID $(\downarrow)$ for Class-Conditional ImageNet $(64\times64)$ dataset by stochastic sampling with NFE = 511.}
  \label{fig:Stochastic_Sampling_ImageNet_FID}
\end{minipage}%
\hfill
\begin{minipage}{.54\textwidth}
  \centering
  \includegraphics[width=1.\linewidth]{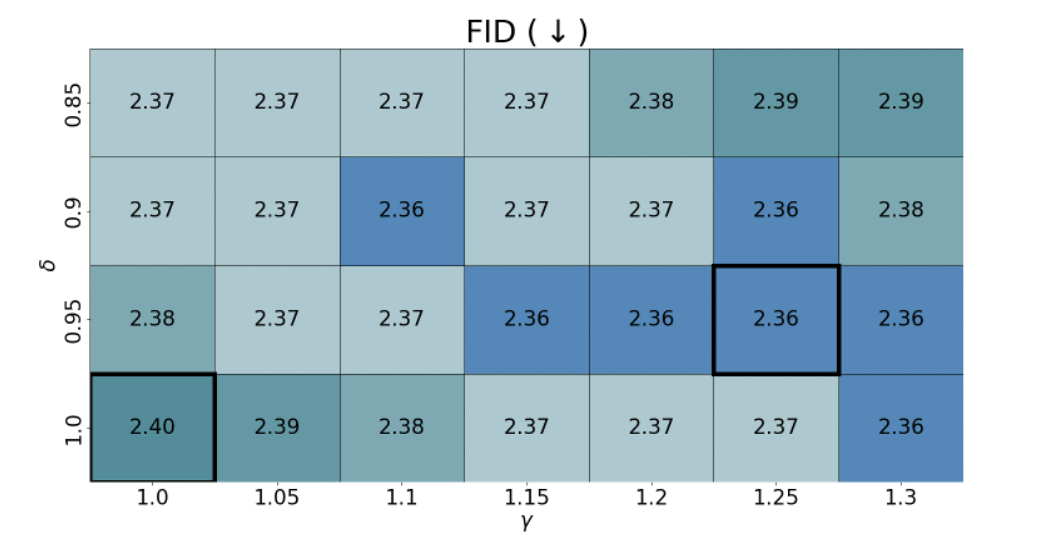}
  \captionof{figure}{Grid search results in FID $(\downarrow)$ for Unconditional CIFAR-10 $(32\times32)$ (VP) by stochastic sampling when varying $\gamma$, $\delta$ with NFE = 511.}
  \label{fig:Stochastic_Sampling_CIFAR10_FID}
\end{minipage}
\end{figure}

We observe synthetic image quality in stochastic sampling $(\rho = 1)$ for numerous values of $\gamma$ and $\delta$. As mentioned before, setting $\gamma = 1$ and $\delta = 1$ allows Corollary \ref{cor:approx_sol} to correspond to the traditional inference formula \ref{eq:kingma_zs}. Similarly to deterministic sampling \ref{exp:deterministic}, different choices of $\gamma$ and $\delta$ can improve results when evaluating model performance. Figure \ref{fig:Stochastic_Sampling_CIFAR10_FID} presents the FID for an unconditional CIFAR-10 $(32 \times 32)$ model from a grid search over $\gamma$ and $\delta$. Among the pool of candidates, the choice of $(\gamma = 1.25, \delta = 0.95)$ yields the best value for this metric and improves model performance beyond $(\gamma = 1, \delta = 1)$, not only in this CIFAR-10 setting but also in an ImageNet $(64 \times 64)$ setting, as shown in Figure \ref{fig:Stochastic_Sampling_ImageNet_FID}.

    

\section{Conclusion}
\label{6_conclusion}
Diffusion models (DMs) have emerged as essential elements within generative models, demonstrating proficiency across diverse domains such as image synthesis, audio generation, and intricate data interpolation. Signal-to-Noise diffusion models encompass a versatile family that includes most cutting-edge diffusion models. While various efforts have been made to analyze Signal-to-Noise (S2N) diffusion models from different angles, there is still a need for a comprehensive investigation that connects disparate perspectives and explores novel viewpoints. In this work, we present an extensive examination of noise schedulers, probing their significance through the prism of the signal-to-noise ratio (SNR) and its links to information theory. Expanding upon this framework, we have devised a generalized backward equation aimed at enhancing the efficacy of the inference process. Our experimental results show that by choosing the correct hyperparameters, our generalized equation improves model performance compared to traditional ones.
\clearpage
\medskip
\bibliographystyle{apalike}
\bibliography{reference}

\clearpage
\appendix
    
\section{Appendix / supplemental material}
\subsection{All Proofs}
\subsubsection{Proof of Theorem \ref{thm:sde}} \label{proof1}

Denote $\Psi\left(\tau,t\right)$ as the transition function satisfying
(i) $\frac{d\Psi\left(\tau,t\right)}{dt}=-\Psi\left(\tau,t\right)f_{t}\mathbf{I}$,
(ii) $\frac{d\Psi\left(\tau,t\right)}{d\tau}=\Psi\left(\tau,t\right)f_{\tau}\mathbf{I}$,
and (iii) $\Psi\left(\tau,\tau\right)=\mathbf{I}$. It is obvious
that we can choose $\Psi\left(\tau,t\right)=\exp\left\{ -\int_{\tau}^{t}f\left(s\right)ds\right\} \mathbf{I}$
that satisfies (i), (ii), and (iii). The distribution $q\left(\boldsymbol{z}_{t}\mid\boldsymbol{z}_{s}\right)=\mathcal{N}\left(m_{t\mid s},\Sigma_{t\mid s}\right)$
is a Gaussian distribution with $m_{t\mid s}=\Psi\left(t,s\right)z_{s}$
and $\Sigma_{t\mid s}=\int_{s}^{t}\Psi\left(t,\tau\right)^{2}g^{2}\left(\tau\right)d\tau$.
Consider $f\left(t\right)=\frac{d\log\alpha\left(t\right)}{dt}=\frac{\alpha'\left(t\right)}{\alpha\left(t\right)}$,
we then have
\[
\Psi\left(\tau,t\right)=\frac{\alpha\left(\tau\right)}{\alpha\left(t\right)}\mathbf{I}.
\]

Therefore, we arrive
\[
m_{t\mid0}=\Psi\left(t,0\right)\boldsymbol{z}_{0}=\alpha\left(t\right)\boldsymbol{z}_{0},
\]
as expected. Moreover, the variance $\Sigma_{t\mid0}$ is
\[
\Sigma_{t\mid0}=\int_{0}^{t}\Psi\left(t,\tau\right)^{2}g^{2}\left(\tau\right)d\tau=\int_{0}^{t}\frac{\alpha\left(t\right)^{2}}{\alpha\left(\tau\right)^{2}}g^{2}\left(\tau\right)d\tau\mathbf{I}.
\]

By choosing $g_{\tau}=\sqrt{-\exp\left\{ -\lambda\left(\tau\right)\right\} \lambda'\left(\tau\right)}\alpha\left(\tau\right)$,
we reach
\[
\Sigma_{t\mid0}=\int_{0}^{t}\frac{\alpha\left(t\right)^{2}}{\alpha\left(\tau\right)^{2}}g^{2}\left(\tau\right)d\tau\mathbf{I}=\alpha^{2}\left(t\right)\exp\left\{ -\lambda\left(t\right)\right\} \mathbf{I}=\sigma^{2}\left(t\right)\mathbf{I}.
\]
as expected. Therefore, The distribution $q\left(z_{t}\mid z_{s}\right)=\mathcal{N}\left(m_{t\mid s},\Sigma_{t\mid s}\right)$
has
\[
m_{t\mid s}=\Psi\left(t,s\right)z_{s}=\frac{\alpha\left(t\right)}{\alpha\left(s\right)}z_{0}=\alpha_{t\mid s}z_{s},
\]
\begin{align*}
\Sigma_{t\mid s} & =\int_{s}^{t}\Psi\left(t,\tau\right)^{2}g^{2}\left(\tau\right)d\tau=\int_{s}^{t}\frac{\alpha\left(t\right)^{2}}{\alpha\left(\tau\right)^{2}}g^{2}\left(\tau\right)d\tau\mathbf{I}=\alpha^{2}\left(t\right)\left[\exp\left\{ -\lambda\left(t\right)\right\} -\exp\left\{ -\lambda\left(s\right)\right\} \right]\\
 & =\alpha^{2}\left(t\right)\left[\frac{1}{SNR\left(t\right)}-\frac{1}{SNR\left(s\right)}\right],
\end{align*}
where we define $SNR\left(t\right)=\frac{\alpha\left(t\right)^{2}}{\sigma\left(t\right)^{2}}.$

The SDE of the forward process is
\[
d\boldsymbol{z}_{t}=\frac{\alpha'(t)}{\alpha\left(t\right)}\boldsymbol{z}_{t}dt+\sqrt{-\exp\left\{ -\lambda\left(t\right)\right\} \lambda'(t)}\alpha\left(t\right)d\boldsymbol{w}_{t}.
\]

\subsubsection{Proof of Theorem \ref{thm:backwardSDE}} \label{proof2}
This proof is similar to that of Prop 1 in \cite{Zhang2022a}.

\clearpage
\subsubsection{Proof of Theorem \ref{thm:backwardSDE_exact}} \label{proof3}

We have
\begin{align*}
dz_{t} & =\left(\frac{\alpha'\left(t\right)}{\alpha\left(t\right)}z_{t}+\frac{1+\lambda^{2}}{2}\exp\left\{ -\lambda\left(t\right)\right\} \lambda'\left(t\right)\alpha^{2}\left(t\right)s_{\theta}\left(z_{t},t\right)\right)dt+\lambda g\left(t\right)d\boldsymbol{w}_{t}\\
 & =\left(\frac{\alpha'\left(t\right)}{\alpha\left(t\right)}z_{t}-\frac{1+\lambda^{2}}{2}\exp\left\{ -\lambda\left(t\right)\right\} \lambda'\left(t\right)\alpha^{2}\left(t\right)\alpha\left(t\right)^{-1}\exp\left\{ \frac{\lambda\left(t\right)}{2}\right\} \epsilon_{\theta}\left(z_{t},t\right)\right)dt+\lambda g\left(t\right)d\boldsymbol{w}_{t}\\
 & =\left(\frac{\alpha'\left(t\right)}{\alpha\left(t\right)}z_{t}-\frac{1+\lambda^{2}}{2}\exp\left\{ \frac{-\lambda\left(t\right)}{2}\right\} \lambda'\left(t\right)\alpha\left(t\right)\epsilon_{\theta}\left(z_{t},t\right)\right)dt+\lambda g\left(t\right)d\boldsymbol{w}_{t}.
\end{align*}

This follows that
\begin{align*}
dz_{\tau}-\frac{\alpha'\left(\tau\right)}{\alpha\left(\tau\right)}z_{\tau} & =-\frac{1+\lambda^{2}}{2}\exp\left\{ \frac{-\lambda\left(\tau\right)}{2}\right\} \lambda'\left(\tau\right)\alpha\left(\tau\right)\epsilon_{\theta}\left(z_{\tau},\tau\right)d\tau\\
 & +\lambda g\left(\tau\right)d\boldsymbol{w}_{\tau}\\
\Psi\left(s,\tau\right)dz_{\tau}-\Psi\left(s,\tau\right)\frac{\alpha'\left(\tau\right)}{\alpha\left(\tau\right)}z_{\tau} & =-\frac{1+\lambda^{2}}{2}\Psi\left(s,\tau\right)\exp\left\{ \frac{-\lambda\left(\tau\right)}{2}\right\} \lambda'\left(\tau\right)\alpha\left(\tau\right)\epsilon_{\theta}\left(z_{\tau},\tau\right)d\tau\\
 & +\lambda\Psi\left(s,\tau\right)g\left(\tau\right)d\boldsymbol{w}_{\tau}\\
\Psi\left(s,\tau\right)dz_{\tau}-\Psi\left(s,\tau\right)f\left(\tau\right)z_{\tau} & =-\frac{1+\lambda^{2}}{2}\Psi\left(s,\tau\right)\exp\left\{ \frac{-\lambda\left(\tau\right)}{2}\right\} \lambda'\left(\tau\right)\alpha\left(\tau\right)\epsilon_{\theta}\left(z_{\tau},\tau\right)d\tau\\
 & +\lambda\Psi\left(s,\tau\right)g\left(\tau\right)d\boldsymbol{w}_{\tau}.
\end{align*}
Using $\nabla_{\tau}\Psi\left(s,\tau\right)=-\Psi\left(s,\tau\right)f\left(\tau\right)z_{\tau}$,
we further reach
\begin{align*}
\Psi\left(s,\tau\right)dz_{\tau}+\nabla_{\tau}\Psi\left(s,\tau\right)z_{\tau} & =-\frac{1+\lambda^{2}}{2}\frac{\alpha\left(s\right)}{\alpha\left(\tau\right)}\exp\left\{ \frac{-\lambda\left(\tau\right)}{2}\right\} \lambda'\left(\tau\right)\alpha\left(\tau\right)\epsilon_{\theta}\left(z_{\tau},\tau\right)d\tau\\
 & +\lambda\frac{\alpha\left(s\right)}{\alpha\left(\tau\right)}g\left(\tau\right)d\boldsymbol{w}_{\tau}\\
d\left(\Psi\left(s,\tau\right)z_{\tau}\right) & =-\frac{1+\lambda^{2}}{2}\alpha\left(s\right)\exp\left\{ \frac{-\lambda\left(\tau\right)}{2}\right\} \lambda'\left(\tau\right)\epsilon_{\theta}\left(z_{\tau},\tau\right)d\tau\\
 & +\lambda\alpha\left(s\right)\sqrt{-\exp\left\{ -\lambda\left(\tau\right)\right\} \lambda'\left(\tau\right)}d\boldsymbol{w}_{\tau}\\
\Psi\left(s,s\right)z_{s}-\Psi\left(s,t\right)z_{t} & =-\frac{1+\lambda^{2}}{2}\alpha\left(s\right)\int_{t}^{s}\exp\left\{ \frac{-\lambda\left(\tau\right)}{2}\right\} \lambda'\left(\tau\right)\epsilon_{\theta}\left(z_{\tau},\tau\right)d\tau\\
 & +\lambda\alpha\left(s\right)\int_{t}^{s}\sqrt{-\exp\left\{ -\lambda\left(\tau\right)\right\} \lambda'\left(\tau\right)}d\boldsymbol{w}_{\tau}.
\end{align*}

\begin{align}
\label{eq:zs}
z_{s}= & \frac{\alpha\left(s\right)}{\alpha\left(t\right)}z_{t}-\frac{1+\lambda^{2}}{2}\alpha\left(s\right)\int_{t}^{s}\exp\left\{ \frac{-\lambda\left(\tau\right)}{2}\right\} \lambda'\left(\tau\right)\epsilon_{\theta}\left(z_{\tau},\tau\right)d\tau\\
 & +\lambda\alpha\left(s\right)\int_{t}^{s}\sqrt{-\exp\left\{ -\lambda\left(\tau\right)\right\} \lambda'\left(\tau\right)}d\boldsymbol{w}_{\tau}.
\end{align}

\subsubsection{Proof of Corollary \ref{cor:approx_sol}}

We first approximate the first regular integral as
\begin{multline}
\label{eq:1st_integral}
\int_{t}^{s}\exp\left\{ \frac{-\lambda\left(\tau\right)}{2}\right\} \lambda'\left(\tau\right)\epsilon_{\theta}\left(z_{\tau},\tau\right)d\tau\\
\approx\int_{t}^{s}\exp\left\{ \frac{-1-\gamma}{2}\lambda\left(\tau\right)\right\} \lambda'\left(\tau\right)d\tau\exp\left\{ \frac{\gamma\lambda\left(t\right)}{2}\right\} \epsilon_{\theta}\left(z_{\tau},t\right)\\
=\int_{t}^{s}\exp\left\{ \frac{-1-\gamma}{2}\lambda\left(\tau\right)\right\} d\lambda\left(\tau\right)\exp\left\{ \frac{\gamma\lambda\left(t\right)}{2}\right\} \epsilon_{\theta}\left(z_{\tau},t\right)\\
=\frac{2}{1+\gamma}\left[\exp\left\{ \frac{-1-\gamma}{2}\lambda\left(t\right)\right\} -\exp\left\{ \frac{-1-\gamma}{2}\lambda\left(s\right)\right\} \right]\exp\left\{ \frac{\gamma\lambda\left(t\right)}{2}\right\} \epsilon_{\theta}\left(z_{\tau},t\right).
\end{multline}

To approximate the second Ito integral, we start from its definition. Consider a partition $P:\tau_{0}=s<\tau_{1}<...<\tau_{n-1}<\tau_{n}=t$,
we have the sum
\begin{align*}
\sum_{i=0}^{n-1}\sqrt{-\exp\left\{ -\lambda\left(\tau_{i}\right)\right\} \lambda'\left(\tau_{i}\right)}\left(\boldsymbol{w}_{i+1}-\boldsymbol{w}_{i}\right) & =\sum_{i=0}^{n-1}\sqrt{-\exp\left\{ -\lambda\left(\tau_{i}\right)\right\} \lambda'\left(\tau_{i}\right)}\sqrt{\tau_{i+1}-\tau_{i}}\boldsymbol{\epsilon_{i}},
\end{align*}

where $\boldsymbol{\epsilon}_{i}\sim\mathcal{N}\left(\boldsymbol{0},\mathbf{I}\right)$.
This further implies that
\begin{align*}
\sum_{i=0}^{n-1}\sqrt{-\exp\left\{ -\lambda\left(\tau_{i}\right)\right\} \lambda'\left(\tau_{i}\right)}\left(\boldsymbol{w}_{i+1}-\boldsymbol{w}_{i}\right) & =\sqrt{-\sum_{i=0}^{n-1}\exp\left\{ -\lambda\left(\tau_{i}\right)\right\} \lambda'\left(\tau_{i}\right)\left(\tau_{i+1}-\tau_{i}\right)}\boldsymbol{\boldsymbol{\epsilon}},
\end{align*}
here we use the property $\sum_{i=0}^{n-1}a_{i}\boldsymbol{\epsilon}_{i}=\sqrt{\sum_{i=0}^{n-1}a_{i}^{2}}\boldsymbol{\epsilon}$
with $\boldsymbol{\epsilon}\sim\mathcal{N}\left(\boldsymbol{0},\mathbf{I}\right)$.

Taking the limit of the above sum when the diameter of the partition
$\delta\left(P\right)=\max_{i}\left(\tau_{i+1}-\tau_{i}\right)$ approaches
$0$, we gain
\[
\lim_{\delta\left(P\right)\rightarrow0}\sum_{i=0}^{n-1}\sqrt{-\exp\left\{ -\lambda\left(\tau_{i}\right)\right\} \lambda'\left(\tau_{i}\right)}\left(\boldsymbol{w}_{i+1}-\boldsymbol{w}_{i}\right)=\sqrt{-\lim_{\delta\left(P\right)\rightarrow0}\sum_{i=0}^{n-1}\exp\left\{ -\lambda\left(\tau_{i}\right)\right\} \lambda'\left(\tau_{i}\right)\left(\tau_{i+1}-\tau_{i}\right)}\boldsymbol{\epsilon}.
\]
\begin{align*}
\int_{s}^{t}\sqrt{-\exp\left\{ -\lambda\left(\tau\right)\right\} \lambda'\left(\tau\right)}d\boldsymbol{w}_{t} & =\sqrt{-\int_{s}^{t}\exp\left\{ -\lambda\left(\tau\right)\right\} \lambda'\left(\tau\right)d\tau}\boldsymbol{\epsilon}\\
 & =\sqrt{\left[\exp\left\{ -\lambda\left(t\right)\right\} -\exp\left\{ -\lambda\left(s\right)\right\} \right]}\boldsymbol{\epsilon.}
\end{align*}
\begin{multline}
\label{eq:2nd_integral}
\alpha\left(s\right)\int_{t}^{s}\sqrt{-\exp\left\{ -\lambda\left(\tau\right)\right\} \lambda'\left(\tau\right)}d\boldsymbol{w}_{t}=\alpha\left(s\right)\sqrt{\left[\exp\left\{ -\lambda\left(t\right)\right\} -\exp\left\{ -\lambda\left(s\right)\right\} \right]}\boldsymbol{\epsilon}\\
\approx\alpha\left(t\right)\sqrt{\left[\exp\left\{ -\lambda\left(t\right)\right\} -\exp\left\{ -\lambda\left(s\right)\right\} \right]}\left(\frac{\alpha\left(s\right)}{\alpha\left(t\right)}\right)^{1-\delta}\left(\frac{\sigma\left(s\right)}{\sigma\left(t\right)}\right)^{\delta}\boldsymbol{\epsilon},
\end{multline}
here we assume that $s=t-\Delta t$ is very close to $t$, hence $\alpha\left(s\right)\approx\alpha\left(t\right)$
and $\sigma\left(s\right)\approx\sigma\left(t\right)$.

Finally, combining Eqs. (\ref{eq:1st_integral}, \ref{eq:2nd_integral}) and Eq. (\ref{eq:zs}), we reach the conclusion. 

\subsubsection{Proof of Theorem \ref{theo:non_Markov}}

We start with
\begin{align*}
z_{s} & =\alpha\left(s\right)\hat{z}_{\theta}\left(z_{t},t\right)+\sqrt{\sigma^{2}\left(s\right)-\beta^{2}\left(s,t\right)}\frac{z_{t}-\alpha\left(t\right)\hat{z}_{\theta}\left(z_{t},t\right)}{\sigma\left(t\right)}+\beta\left(s\right)\boldsymbol{\epsilon}\\
 & =\left(\alpha\left(s\right)-\frac{\alpha\left(t\right)\sqrt{\sigma^{2}\left(s\right)-\beta^{2}\left(s,t\right)}}{\sigma\left(t\right)}\right)\hat{z}_{\theta}\left(z_{t},t\right)+\frac{\sqrt{\sigma^{2}\left(s\right)-\beta^{2}\left(s,t\right)}}{\sigma\left(t\right)}z_{t}+\beta\left(s,t\right)\boldsymbol{\epsilon}\\
 & =\left(\alpha\left(s\right)-\frac{\alpha\left(t\right)\sqrt{\sigma^{2}\left(s\right)-\beta^{2}\left(s,t\right)}}{\sigma\left(t\right)}\right)\frac{z_{t}-\sigma\left(t\right)\epsilon_{\theta}\left(z_{t},t\right)}{\alpha\left(t\right)}+\frac{\sqrt{\sigma^{2}\left(s\right)-\beta^{2}\left(s,t\right)}}{\sigma\left(t\right)}z_{t}+\beta\left(s,t\right)\boldsymbol{\epsilon}\\
 & =-\left(\alpha\left(s\right)-\frac{\alpha\left(t\right)\sqrt{\sigma^{2}\left(s\right)-\beta^{2}\left(s,t\right)}}{\sigma\left(t\right)}\right)\frac{\sigma\left(t\right)\epsilon_{\theta}\left(z_{t},t\right)}{\alpha\left(t\right)}\\
 & +\left(\frac{\alpha\left(s\right)}{\alpha\left(t\right)}+\frac{\sqrt{\sigma^{2}\left(s\right)-\beta^{2}\left(s,t\right)}}{\sigma\left(t\right)}\right)z_{t}+\beta\left(s,t\right)\boldsymbol{\epsilon}\\
 & =-\left(\frac{\alpha\left(s\right)}{\sigma\left(s\right)}-\frac{\alpha\left(t\right)\sqrt{1-\frac{\beta^{2}\left(s,t\right)}{\sigma^{2}\left(s\right)}}}{\sigma\left(t\right)}\right)\frac{\sigma\left(t\right)\sigma\left(s\right)\epsilon_{\theta}\left(z_{t},t\right)}{\alpha\left(t\right)}+\left(\frac{\alpha\left(s\right)}{\alpha\left(t\right)}+\frac{\sigma\left(s\right)\sqrt{1-\frac{\beta^{2}\left(s,t\right)}{\sigma^{2}\left(s\right)}}}{\sigma\left(t\right)}\right)z_{t}+\beta\left(s,t\right)\boldsymbol{\epsilon}\\
 & =-\left(\exp\left\{ \frac{\lambda\left(s\right)}{2}\right\} -\exp\left\{ \frac{\lambda\left(t\right)}{2}\right\} \sqrt{1-\frac{\beta^{2}\left(s,t\right)}{\sigma^{2}\left(s\right)}}\right)\exp\left\{ \frac{-\lambda\left(t\right)-\lambda\left(s\right)}{2}\right\} \alpha\left(s\right)\epsilon_{\theta}\left(z_{t},t\right)\\
+ & \frac{\alpha\left(s\right)}{\alpha\left(t\right)}\left(1+\exp\left\{ \frac{\lambda\left(t\right)-\lambda\left(s\right)}{2}\right\} \sqrt{1-\frac{\beta^{2}\left(s,t\right)}{\sigma^{2}\left(s\right)}}\right)z_{t}+\beta\left(s,t\right)\boldsymbol{\epsilon}.
\end{align*}

Let $s= t- \Delta t$, we gain
\begin{align*}
\boldsymbol{z}_{t-\Delta t} & =-\left(\exp\left\{ \frac{\lambda\left(t-\Delta t\right)}{2}\right\} -\exp\left\{ \frac{\lambda\left(t\right)}{2}\right\} \sqrt{1-\frac{\beta^{2}\left(t-\Delta t,t\right)}{\sigma^{2}\left(t-\Delta t\right)}}\right)\exp\left\{ \frac{-\lambda\left(t\right)-\lambda\left(t-\Delta t\right)}{2}\right\} \alpha\left(t-\Delta t\right)\epsilon_{\theta}\left(z_{t},t\right)\\
 & +\frac{\alpha\left(t-\Delta t\right)}{\alpha\left(t\right)}\left(1+\exp\left\{ \frac{\lambda\left(t\right)-\lambda\left(t-\Delta t\right)}{2}\right\} \sqrt{1-\frac{\beta^{2}\left(t-\Delta t,t\right)}{\sigma^{2}\left(t-\Delta t\right)}}\right)z_{t}+\beta\left(t-\Delta t,t\right)\epsilon
\end{align*}
\begin{multline*}
\frac{\boldsymbol{z}_{t-\Delta t}-\boldsymbol{z}_{t}}{-\Delta t}=-\left(\frac{\exp\left\{ \frac{\lambda\left(t-\Delta t\right)}{2}\right\} -\exp\left\{ \frac{\lambda\left(t\right)}{2}\right\} \sqrt{1-\frac{\beta^{2}\left(t-\Delta t\right)}{\sigma^{2}\left(t-\Delta t\right)}}}{-\Delta t}\right)\exp\left\{ \frac{-\lambda\left(t\right)-\lambda\left(t-\Delta t\right)}{2}\right\} \alpha\left(t-\Delta t\right)\epsilon_{\theta}\left(z_{t},t\right)\\
+\frac{\alpha\left(t-\Delta t\right)-\alpha\left(t\right)}{-\Delta t\alpha\left(t\right)}\boldsymbol{z}_{t}+\frac{\alpha\left(t-\Delta t\right)}{\alpha\left(t\right)}\frac{\exp\left\{ \frac{\lambda\left(t\right)-\lambda\left(t-\Delta t\right)}{2}\right\} }{-\Delta t}\sqrt{1-\frac{\beta^{2}\left(t-\Delta t\right)}{\sigma^{2}\left(t-\Delta t\right)}}z_{t}+\frac{\beta\left(t-\Delta t\right)}{\sqrt{\Delta t}}\frac{\boldsymbol{w}_{t-\Delta t}-\boldsymbol{w}_{t}}{-\Delta t}
\end{multline*}
This follows that
\begin{multline*}
\frac{\boldsymbol{z}_{t-\Delta t}-\boldsymbol{z}_{t}}{-\Delta t}=-\frac{\exp\left\{ \frac{\lambda\left(t-\Delta t\right)}{2}\right\} -\exp\left\{ \frac{\lambda\left(t\right)}{2}\right\} }{-\Delta t}\exp\left\{ \frac{-\lambda\left(t\right)-\lambda\left(t-\Delta t\right)}{2}\right\} \alpha\left(t-\Delta t\right)\epsilon_{\theta}\left(z_{t},t\right)\\
+\exp\left\{ \frac{\lambda\left(t\right)}{2}\right\} \frac{\beta^{2}\left(t-\Delta t\right)}{-\Delta t\sigma^{2}\left(t-\Delta t\right)}\frac{1}{1+\sqrt{1-\frac{\beta^{2}\left(t-\Delta t\right)}{\sigma^{2}\left(t-\Delta t\right)}}}\exp\left\{ \frac{-\lambda\left(t\right)-\lambda\left(t-\Delta t\right)}{2}\right\} \alpha\left(t-\Delta t\right)\epsilon_{\theta}\left(z_{t},t\right)\\
+\frac{\alpha\left(t-\Delta t\right)-\alpha\left(t\right)}{-\Delta t\alpha\left(t\right)}\boldsymbol{z}_{t}+\frac{\alpha\left(t-\Delta t\right)}{\alpha\left(t\right)}\frac{\exp\left\{ \frac{\lambda\left(t\right)-\lambda\left(t-\Delta t\right)}{2}\right\} }{-\Delta t}\sqrt{1-\frac{\beta^{2}\left(t-\Delta t\right)}{\sigma^{2}\left(t-\Delta t\right)}}z_{t}\\
+\frac{\beta\left(t-\Delta t\right)}{\sqrt{\Delta t}}\frac{\boldsymbol{w}_{t-\Delta t}-\boldsymbol{w}_{t}}{-\Delta t}
\end{multline*}

Let consider $\beta\left(s,t\right)=\sqrt{b\left(s\right)-b\left(t\right)}$,
we have
\begin{align*}
\frac{\boldsymbol{z}_{t-\Delta t}-\boldsymbol{z}_{t}}{-\Delta t} & =-\frac{\exp\left\{ \frac{\lambda\left(t-\Delta t\right)}{2}\right\} -\exp\left\{ \frac{\lambda\left(t\right)}{2}\right\} }{-\Delta t}\exp\left\{ \frac{-\lambda\left(t\right)-\lambda\left(t-\Delta t\right)}{2}\right\} \alpha\left(t-\Delta t\right)\epsilon_{\theta}\left(z_{t},t\right)\\
+ & \exp\left\{ \frac{\lambda\left(t\right)}{2}\right\} \frac{b\left(t-\Delta t\right)-b\left(t\right)}{-\Delta t\sigma^{2}\left(t-\Delta t\right)}\frac{1}{1+\sqrt{1-\frac{b\left(t-\Delta t\right)-b\left(t\right)}{\sigma^{2}\left(t-\Delta t\right)}}}\exp\left\{ \frac{-\lambda\left(t\right)-\lambda\left(t-\Delta t\right)}{2}\right\} \alpha\left(t-\Delta t\right)\epsilon_{\theta}\left(z_{t},t\right)\\
+ & \frac{\alpha\left(t-\Delta t\right)-\alpha\left(t\right)}{-\Delta t\alpha\left(t\right)}\boldsymbol{z}_{t}+\frac{\alpha\left(t-\Delta t\right)}{\alpha\left(t\right)}\frac{\exp\left\{ \frac{\lambda\left(t\right)-\lambda\left(t-\Delta t\right)}{2}\right\} }{-\Delta t}\sqrt{1-\frac{b\left(t-\Delta t\right)-b\left(t\right)}{\sigma^{2}\left(t-\Delta t\right)}}z_{t}\\
 & +\frac{\sqrt{b\left(t-\Delta t\right)-b\left(t\right)}}{\sqrt{\Delta t}}\frac{\boldsymbol{w}_{t-\Delta t}-\boldsymbol{w}_{t}}{-\Delta t}
\end{align*}

\begin{align}
d\boldsymbol{z}_{t} & =\left[\frac{\alpha'\left(t\right)}{\alpha\left(t\right)}+\frac{\lambda'\left(t\right)}{2}\exp\left\{ \frac{\lambda\left(t\right)}{2}\right\} \right]\boldsymbol{\boldsymbol{z}}_{t}-\frac{\lambda'\left(t\right)}{2}\exp\left\{ -\frac{\lambda\left(t\right)}{2}\right\} \alpha\left(t\right)\epsilon_{\theta}\left(\boldsymbol{z}_{t},t\right)dt\nonumber \\
 & +\frac{1}{2}b'\left(t\right)\exp\left\{ \frac{\lambda\left(t\right)}{2}\right\} \alpha^{-1}\left(t\right)\epsilon_{\theta}\left(z_{t},t\right)+\sqrt{-b'\left(t\right)}d\boldsymbol{w}_{t}.\label{eq:non_markov_SDE_proof}
\end{align}

\subsubsection{Proof of Theorem \ref{thm:equivalent}}

The proof is quite obvious from $\tilde{\alpha}_{1}=\alpha_{1}\circ\lambda_{1}^{-1}=\alpha_{2}\circ\lambda_{2}^{-1}=\tilde{\alpha}_{2}$ and $\tilde{\sigma}_{1}=\sigma_{1}\circ\lambda_{1}^{-1}=\sigma_{2}\circ\lambda_{2}^{-1}=\tilde{\sigma}_{2}$.

\subsubsection{Proof of Theorem \ref{theo:information}}

We start with
\begin{align*}
\frac{d}{d\lambda}D_{KL}\left(p\left(\tilde{\boldsymbol{z}}_{\lambda}\mid\boldsymbol{x}\right)\Vert p\left(\tilde{\boldsymbol{z}}_{\lambda}\right)\right) & =\frac{d}{d\lambda}\mathbb{E}_{p\left(\tilde{\boldsymbol{z}}_{\lambda}\mid\boldsymbol{x}\right)}\left[\log p\left(\tilde{\boldsymbol{z}}_{\lambda}\mid\boldsymbol{x}\right)\right]-\frac{d}{d\lambda}\mathbb{E}_{p\left(\tilde{\boldsymbol{z}}_{\lambda}\mid\boldsymbol{x}\right)}\left[\log p\left(\tilde{\boldsymbol{z}}_{\lambda}\right)\right]\\
 & =A-B
\end{align*}
\begin{align*}
A=\frac{d}{d\lambda}\mathbb{E}_{p\left(\tilde{\boldsymbol{z}}_{\lambda}\mid\boldsymbol{x}\right)}\left[\log p\left(\tilde{\boldsymbol{z}}_{\lambda}\mid\boldsymbol{x}\right)\right] & =-\frac{1}{2}\frac{d}{d\lambda}\log\left[\tilde{\sigma}\left(\lambda\right)\right]^{D}=-\frac{D\tilde{\sigma}'\left(\lambda\right)}{2\tilde{\sigma}\left(\lambda\right)}
\end{align*}
\begin{align*}
B=\frac{d}{d\lambda}\mathbb{E}_{p\left(\tilde{\boldsymbol{z}}_{\lambda}\mid\boldsymbol{x}\right)}\left[\log p\left(\tilde{\boldsymbol{z}}_{\lambda}\right)\right] & =\frac{d}{d\lambda}\int\log p\left(\tilde{\boldsymbol{z}}_{\lambda}\right)p\left(\tilde{\boldsymbol{z}}_{\lambda}\mid\boldsymbol{x}\right)d\tilde{\boldsymbol{z}}_{\lambda}\\
= & \int\left[p\left(\tilde{\boldsymbol{z}}_{\lambda}\mid\boldsymbol{x}\right)\frac{d}{d\lambda}\log p\left(\tilde{\boldsymbol{z}}_{\lambda}\right)+\log p\left(\tilde{\boldsymbol{z}}_{\lambda}\right)\frac{d}{d\lambda}p\left(\tilde{\boldsymbol{z}}_{\lambda}\mid\boldsymbol{x}\right)\right]d\tilde{\boldsymbol{z}}_{\lambda}.
\end{align*}
\begin{align*}
\frac{d}{d\lambda}p\left(\tilde{\boldsymbol{z}}_{\lambda}\mid\boldsymbol{x}\right) & =\nabla_{\lambda}\frac{-\tilde{\alpha}\left(\lambda\right)\boldsymbol{x}}{\tilde{\sigma}\left(\lambda\right)}\cdot\nabla_{\tilde{\boldsymbol{z}}_{\lambda}}p\left(\tilde{\boldsymbol{z}}_{\lambda}\mid\boldsymbol{x}\right)\\
 & =-\frac{\tilde{\alpha}'\left(\lambda\right)\tilde{\sigma}\left(\lambda\right)-\tilde{\alpha}\left(\lambda\right)\tilde{\sigma}'\left(\lambda\right)}{\tilde{\sigma}^{2}\left(\lambda\right)}\boldsymbol{x}\cdot\nabla_{\tilde{\boldsymbol{z}}_{\lambda}}p\left(\tilde{\boldsymbol{z}}_{\lambda}\mid\boldsymbol{x}\right)\\
 & =c\left(\lambda\right)\boldsymbol{x}\cdot\nabla_{\tilde{\boldsymbol{z}}_{\lambda}}p\left(\tilde{\boldsymbol{z}}_{\lambda}\mid\boldsymbol{x}\right),
\end{align*}
where $c\left(\lambda\right)=-\frac{\tilde{\alpha}'\left(\lambda\right)\tilde{\sigma}\left(\lambda\right)-\tilde{\alpha}\left(\lambda\right)\tilde{\sigma}'\left(\lambda\right)}{\tilde{\sigma}^{2}\left(\lambda\right)}$.
\begin{align*}
\frac{d}{d\lambda}\log p\left(\tilde{\boldsymbol{z}}_{\lambda}\right) & =\frac{1}{p\left(\tilde{\boldsymbol{z}}_{\lambda}\right)}\frac{d}{d\lambda}p\left(\tilde{\boldsymbol{z}}_{\lambda}\right)=\frac{1}{p\left(\tilde{\boldsymbol{z}}_{\lambda}\right)}\frac{d}{d\lambda}\int p\left(\tilde{\boldsymbol{z}}_{\lambda}\mid\boldsymbol{\bar{x}}\right)p\left(\bar{\boldsymbol{x}}\right)d\bar{\boldsymbol{x}}\\
= & \frac{1}{p\left(\tilde{\boldsymbol{z}}_{\lambda}\right)}\int c\left(\lambda\right)\bar{\boldsymbol{x}}\cdot\nabla_{\tilde{\boldsymbol{z}}_{\lambda}}p\left(\tilde{\boldsymbol{z}}_{\lambda}\mid\boldsymbol{\bar{x}}\right)p\left(\bar{\boldsymbol{x}}\right)d\bar{\boldsymbol{x}}.
\end{align*}

Using integral-by-part, we gain

\begin{align*}
\int\log p\left(\tilde{\boldsymbol{z}}_{\lambda}\right)\frac{d}{d\lambda}p\left(\tilde{\boldsymbol{z}}_{\lambda}\mid\boldsymbol{x}\right)d\tilde{\boldsymbol{z}}_{\lambda} & =-c\left(\lambda\right)\boldsymbol{x}\cdot\int\log p\left(\tilde{\boldsymbol{z}}_{\lambda}\right)\nabla_{\tilde{\boldsymbol{z}}_{\lambda}}p\left(\tilde{\boldsymbol{z}}_{\lambda}\mid\boldsymbol{x}\right)d\tilde{\boldsymbol{z}}_{\lambda}\\
= & -c\left(\lambda\right)\boldsymbol{x}\cdot\int p\left(\tilde{\boldsymbol{z}}_{\lambda}\mid\boldsymbol{x}\right)\nabla_{\tilde{\boldsymbol{z}}_{\lambda}}\log p\left(\tilde{\boldsymbol{z}}_{\lambda}\right)d\tilde{\boldsymbol{z}}_{\lambda}.
\end{align*}
\begin{align*}
\int p\left(\tilde{\boldsymbol{z}}_{\lambda}\mid\boldsymbol{x}\right)\frac{d}{d\lambda}\log p\left(\tilde{\boldsymbol{z}}_{\lambda}\right)d\tilde{\boldsymbol{z}}_{\lambda} & =\int\int\frac{p\left(\tilde{\boldsymbol{z}}_{\lambda}\mid\boldsymbol{x}\right)}{p\left(\tilde{\boldsymbol{z}}_{\lambda}\right)}c\left(\lambda\right)\bar{\boldsymbol{x}}\cdot\nabla_{\tilde{\boldsymbol{z}}_{\lambda}}p\left(\tilde{\boldsymbol{z}}_{\lambda}\mid\boldsymbol{\bar{x}}\right)p\left(\bar{\boldsymbol{x}}\right)d\tilde{\boldsymbol{z}}_{\lambda}d\bar{\boldsymbol{x}}\\
=\int & \int\frac{p\left(\boldsymbol{x}\mid\tilde{\boldsymbol{z}}_{\lambda}\right)}{p\left(\boldsymbol{x}\right)}c\left(\lambda\right)\bar{\boldsymbol{x}}\cdot\nabla_{\tilde{\boldsymbol{z}}_{\lambda}}p\left(\tilde{\boldsymbol{z}}_{\lambda}\mid\boldsymbol{\bar{x}}\right)p\left(\bar{\boldsymbol{x}}\right)d\tilde{\boldsymbol{z}}_{\lambda}d\bar{\boldsymbol{x}}\\
=\frac{1}{p\left(\boldsymbol{x}\right)}c\left(\lambda\right) & \int p\left(\bar{\boldsymbol{x}}\right)\bar{\boldsymbol{x}}\cdot\int p\left(\boldsymbol{x}\mid\tilde{\boldsymbol{z}}_{\lambda}\right)\nabla_{\tilde{\boldsymbol{z}}_{\lambda}}p\left(\tilde{\boldsymbol{z}}_{\lambda}\mid\boldsymbol{\bar{x}}\right)d\tilde{\boldsymbol{z}}_{\lambda}d\bar{\boldsymbol{x}}\\
=- & \frac{1}{p\left(\boldsymbol{x}\right)}c\left(\lambda\right)\int\bar{\boldsymbol{x}}\cdot\nabla_{\tilde{\boldsymbol{z}}_{\lambda}}p\left(\boldsymbol{x}\mid\tilde{\boldsymbol{z}}_{\lambda}\right)p\left(\tilde{\boldsymbol{z}}_{\lambda}\mid\boldsymbol{\bar{x}}\right)p\left(\bar{\boldsymbol{x}}\right)d\bar{\boldsymbol{x}}d\tilde{\boldsymbol{z}}_{\lambda}\\
= & -\frac{1}{p\left(\boldsymbol{x}\right)}c\left(\lambda\right)\int\int\bar{\boldsymbol{x}}\cdot p\left(\boldsymbol{\bar{x}},\tilde{\boldsymbol{z}}_{\lambda}\right)d\bar{\boldsymbol{x}}\cdot\nabla_{\tilde{\boldsymbol{z}}_{\lambda}}p\left(\boldsymbol{x}\mid\tilde{\boldsymbol{z}}_{\lambda}\right)d\tilde{\boldsymbol{z}}_{\lambda}\\
= & -\frac{1}{p\left(\boldsymbol{x}\right)}c\left(\lambda\right)\int\mathbb{E}\left[\boldsymbol{x}\mid\tilde{\boldsymbol{z}}_{\lambda}\right]\cdot\nabla_{\tilde{\boldsymbol{z}}_{\lambda}}p\left(\boldsymbol{x}\mid\tilde{\boldsymbol{z}}_{\lambda}\right)p\left(\tilde{\boldsymbol{z}}_{\lambda}\right)d\tilde{\boldsymbol{z}}_{\lambda}.
\end{align*}
\begin{align*}
B & =-c\left(\lambda\right)\boldsymbol{x}\cdot\int p\left(\tilde{\boldsymbol{z}}_{\lambda}\mid\boldsymbol{x}\right)\nabla_{\tilde{\boldsymbol{z}}_{\lambda}}\log p\left(\tilde{\boldsymbol{z}}_{\lambda}\right)d\tilde{\boldsymbol{z}}_{\lambda}\\
 & +\frac{1}{p\left(\boldsymbol{x}\right)}c\left(\lambda\right)\int\mathbb{E}\left[\boldsymbol{x}\mid\tilde{\boldsymbol{z}}_{\lambda}\right]\cdot\nabla_{\tilde{\boldsymbol{z}}_{\lambda}}p\left(\boldsymbol{x}\mid\tilde{\boldsymbol{z}}_{\lambda}\right)p\left(\tilde{\boldsymbol{z}}_{\lambda}\right)d\tilde{\boldsymbol{z}}_{\lambda}.
\end{align*}
\begin{align*}
\nabla_{\tilde{\boldsymbol{z}}_{\lambda}}\log p\left(\tilde{\boldsymbol{z}}_{\lambda}\right) & =\frac{\nabla_{\tilde{\boldsymbol{z}}_{\lambda}}p\left(\tilde{\boldsymbol{z}}_{\lambda}\right)}{p\left(\tilde{\boldsymbol{z}}_{\lambda}\right)}=\frac{\nabla_{\tilde{\boldsymbol{z}}_{\lambda}}\int p\left(\tilde{\boldsymbol{z}}_{\lambda}\mid\bar{\boldsymbol{x}}\right)p\left(\bar{\boldsymbol{x}}\right)d\bar{\boldsymbol{x}}}{p\left(\tilde{\boldsymbol{z}}_{\lambda}\right)}=\frac{\int\nabla_{\tilde{\boldsymbol{z}}_{\lambda}}p\left(\tilde{\boldsymbol{z}}_{\lambda}\mid\bar{\boldsymbol{x}}\right)p\left(\bar{\boldsymbol{x}}\right)d\bar{\boldsymbol{x}}}{p\left(\tilde{\boldsymbol{z}}_{\lambda}\right)}\\
 & =\frac{\int\frac{\tilde{\boldsymbol{z}}_{\lambda}-\tilde{\alpha}\left(\lambda\right)\bar{\boldsymbol{x}}}{\tilde{\sigma}\left(\lambda\right)}p\left(\tilde{\boldsymbol{z}}_{\lambda}\mid\bar{\boldsymbol{x}}\right)p\left(\bar{\boldsymbol{x}}\right)d\bar{\boldsymbol{x}}}{p\left(\tilde{\boldsymbol{z}}_{\lambda}\right)}=\int\frac{\tilde{\boldsymbol{z}}_{\lambda}-\tilde{\alpha}\left(\lambda\right)\bar{\boldsymbol{x}}}{\tilde{\sigma}\left(\lambda\right)}p\left(\bar{\boldsymbol{x}}\mid\tilde{\boldsymbol{z}}_{\lambda}\right)d\bar{\boldsymbol{x}}\\
 & =-\frac{\tilde{\alpha}\left(\lambda\right)}{\tilde{\sigma}\left(\lambda\right)}\mathbb{E}\left[\bar{\boldsymbol{x}}\mid\tilde{\boldsymbol{z}}_{\lambda}\right]+\frac{1}{\tilde{\sigma}\left(\lambda\right)}\tilde{\boldsymbol{z}}_{\lambda}=-\frac{\tilde{\alpha}\left(\lambda\right)}{\tilde{\sigma}\left(\lambda\right)}\mathbb{E}\left[\boldsymbol{x}\mid\tilde{\boldsymbol{z}}_{\lambda}\right]+\frac{1}{\tilde{\sigma}\left(\lambda\right)}\tilde{\boldsymbol{z}}_{\lambda}.
\end{align*}
\begin{align*}
\nabla_{\tilde{\boldsymbol{z}}_{\lambda}}\log p\left(\boldsymbol{x}\mid\tilde{\boldsymbol{z}}_{\lambda}\right) & =\nabla_{\tilde{\boldsymbol{z}}_{\lambda}}\log p\left(\tilde{\boldsymbol{z}}_{\lambda}\mid\boldsymbol{x}\right)-\nabla_{\tilde{\boldsymbol{z}}_{\lambda}}\log p\left(\tilde{\boldsymbol{z}}_{\lambda}\right)\\
= & \frac{\tilde{\boldsymbol{z}}_{\lambda}-\tilde{\alpha}\left(\lambda\right)\boldsymbol{x}}{\tilde{\sigma}\left(\lambda\right)}-\left[-\frac{\tilde{\alpha}\left(\lambda\right)}{\tilde{\sigma}\left(\lambda\right)}\mathbb{E}\left[\boldsymbol{x}\mid\tilde{\boldsymbol{z}}_{\lambda}\right]+\frac{1}{\tilde{\sigma}\left(\lambda\right)}\tilde{\boldsymbol{z}}_{\lambda}\right]\\
= & -\frac{\tilde{\alpha}\left(\lambda\right)}{\tilde{\sigma}\left(\lambda\right)}\left[\boldsymbol{x}-\mathbb{E}\left[\boldsymbol{x}\mid\tilde{\boldsymbol{z}}_{\lambda}\right]\right].
\end{align*}
\[
\nabla_{\tilde{\boldsymbol{z}}_{\lambda}}p\left(\boldsymbol{x}\mid\tilde{\boldsymbol{z}}_{\lambda}\right)=p\left(\boldsymbol{x}\mid\tilde{\boldsymbol{z}}_{\lambda}\right)\nabla_{\tilde{\boldsymbol{z}}_{\lambda}}\log p\left(\boldsymbol{x}\mid\tilde{\boldsymbol{z}}_{\lambda}\right)=-\frac{\tilde{\alpha}\left(\lambda\right)}{\tilde{\sigma}\left(\lambda\right)}\left[\boldsymbol{x}-\mathbb{E}\left[\bar{\boldsymbol{x}}\mid\tilde{\boldsymbol{z}}_{\lambda}\right]\right]p\left(\boldsymbol{x}\mid\tilde{\boldsymbol{z}}_{\lambda}\right).
\]
\begin{align*}
B= & c\left(\lambda\right)\boldsymbol{x}\cdot\int p\left(\tilde{\boldsymbol{z}}_{\lambda}\mid\boldsymbol{x}\right)\left[\frac{\tilde{\alpha}\left(\lambda\right)}{\tilde{\sigma}\left(\lambda\right)}\mathbb{E}\left[\boldsymbol{x}\mid\tilde{\boldsymbol{z}}_{\lambda}\right]-\frac{1}{\tilde{\sigma}\left(\lambda\right)}\tilde{\boldsymbol{z}}_{\lambda}\right]d\tilde{\boldsymbol{z}}_{\lambda}\\
 & -\frac{1}{p\left(\boldsymbol{x}\right)}c\left(\lambda\right)\int\mathbb{E}\left[\boldsymbol{x}\mid\tilde{\boldsymbol{z}}_{\lambda}\right]\cdot\frac{\tilde{\alpha}\left(\lambda\right)}{\tilde{\sigma}\left(\lambda\right)}\left[\boldsymbol{x}-\mathbb{E}\left[\boldsymbol{x}\mid\tilde{\boldsymbol{z}}_{\lambda}\right]\right]p\left(\boldsymbol{x}\mid\tilde{\boldsymbol{z}}_{\lambda}\right)p\left(\tilde{\boldsymbol{z}}_{\lambda}\right)d\tilde{\boldsymbol{z}}_{\lambda}\\
= & \frac{c\left(\lambda\right)}{\tilde{\sigma}\left(\lambda\right)}\int\left[\tilde{\alpha}\left(\lambda\right)\boldsymbol{x}\cdot\mathbb{E}\left[\boldsymbol{x}\mid\tilde{\boldsymbol{z}}_{\lambda}\right]-\boldsymbol{x}\cdot\tilde{\boldsymbol{z}}_{\lambda}\right]p\left(\tilde{\boldsymbol{z}}_{\lambda}\mid\boldsymbol{x}\right)d\tilde{\boldsymbol{z}}_{\lambda}\\
 & -\frac{c\left(\lambda\right)\tilde{\alpha}\left(\lambda\right)}{\tilde{\sigma}\left(\lambda\right)}\int\mathbb{E}\left[\boldsymbol{x}\mid\tilde{\boldsymbol{z}}_{\lambda}\right]\cdot\left[\boldsymbol{x}-\mathbb{E}\left[\boldsymbol{x}\mid\tilde{\boldsymbol{z}}_{\lambda}\right]\right]p\left(\tilde{\boldsymbol{z}}_{\lambda}\mid\boldsymbol{x}\right)d\tilde{\boldsymbol{z}}_{\lambda}\\
= & -\frac{c\left(\lambda\right)}{\tilde{\sigma}\left(\lambda\right)}\int\left[\tilde{\alpha}\left(\lambda\right)\boldsymbol{x}\cdot\boldsymbol{x}-\tilde{\alpha}\left(\lambda\right)\boldsymbol{x}\cdot\mathbb{E}\left[\boldsymbol{x}\mid\tilde{\boldsymbol{z}}_{\lambda}\right]\right]p\left(\tilde{\boldsymbol{z}}_{\lambda}\mid\boldsymbol{x}\right)d\tilde{\boldsymbol{z}}_{\lambda}\\
 & -\frac{c\left(\lambda\right)\tilde{\alpha}\left(\lambda\right)}{\tilde{\sigma}\left(\lambda\right)}\int\mathbb{E}\left[\boldsymbol{x}\mid\tilde{\boldsymbol{z}}_{\lambda}\right]\cdot\left[\boldsymbol{x}-\mathbb{E}\left[\boldsymbol{x}\mid\tilde{\boldsymbol{z}}_{\lambda}\right]\right]p\left(\tilde{\boldsymbol{z}}_{\lambda}\mid\boldsymbol{x}\right)d\tilde{\boldsymbol{z}}_{\lambda}\\
= & -\frac{c\left(\lambda\right)\tilde{\alpha}\left(\lambda\right)}{\tilde{\sigma}\left(\lambda\right)}\int\Vert\boldsymbol{x}-\mathbb{E}\left[\boldsymbol{x}\mid\tilde{\boldsymbol{z}}_{\lambda}\right]\Vert^{2}p\left(\tilde{\boldsymbol{z}}_{\lambda}\mid\boldsymbol{x}\right)d\tilde{\boldsymbol{z}}_{\lambda}=\frac{\left[\tilde{\alpha}'\left(\lambda\right)\tilde{\sigma}\left(\lambda\right)-\tilde{\alpha}\left(\lambda\right)\tilde{\sigma}'\left(\lambda\right)\right]\tilde{\alpha}\left(\lambda\right)}{\tilde{\sigma}^{3}\left(\lambda\right)}\text{mmse}\left(\boldsymbol{x},\lambda\right).
\end{align*}

Therefore, we reach
\[
\frac{d}{d\lambda}D_{KL}\left(p\left(\tilde{\boldsymbol{z}}_{\lambda}\mid\boldsymbol{x}\right)\Vert p\left(\tilde{\boldsymbol{z}}_{\lambda}\right)\right)=-\frac{D\tilde{\sigma}'\left(\lambda\right)}{2\tilde{\sigma}\left(\lambda\right)}+\frac{\left[\tilde{\alpha}'\left(\lambda\right)\tilde{\sigma}\left(\lambda\right)-\tilde{\alpha}\left(\lambda\right)\tilde{\sigma}'\left(\lambda\right)\right]\tilde{\alpha}\left(\lambda\right)}{\tilde{\sigma}^{3}\left(\lambda\right)}\text{mmse}\left(\boldsymbol{x},\lambda\right)
\]

This further implies that
\begin{align*}
\frac{d}{d\lambda}\mathbb{I}\left(\boldsymbol{x},\boldsymbol{\tilde{z}}_{\lambda}\right) & =\frac{d}{d\lambda}\mathbb{E}_{\boldsymbol{x}}\left[D_{KL}\left(p\left(\tilde{\boldsymbol{z}}_{\lambda}\mid\boldsymbol{x}\right)\Vert p\left(\tilde{\boldsymbol{z}}_{\lambda}\right)\right)\right]\\
 & =-\frac{D\tilde{\sigma}'\left(\lambda\right)}{2\tilde{\sigma}\left(\lambda\right)}+\frac{\left[\tilde{\alpha}'\left(\lambda\right)\tilde{\sigma}\left(\lambda\right)-\tilde{\alpha}\left(\lambda\right)\tilde{\sigma}'\left(\lambda\right)\right]\tilde{\alpha}\left(\lambda\right)}{\tilde{\sigma}^{3}\left(\lambda\right)}\text{mmse}\left(\lambda\right)
\end{align*}
\[
\frac{d}{d\lambda}\mathbb{I}\left(\boldsymbol{x},\boldsymbol{\tilde{z}}_{\lambda}\right)=-\frac{D\tilde{\sigma}'\left(\lambda\right)}{2\tilde{\sigma}\left(\lambda\right)}+\frac{\left[\tilde{\alpha}'\left(\lambda\right)\tilde{\sigma}\left(\lambda\right)-\tilde{\alpha}\left(\lambda\right)\tilde{\sigma}'\left(\lambda\right)\right]\tilde{\alpha}\left(\lambda\right)}{\tilde{\sigma}^{3}\left(\lambda\right)}\text{mmse}\left(\lambda\right)
\]

\clearpage
\subsection{Additional Visualizations}

\begin{figure*}[ht]
\centering
    \includegraphics[width=1.0\textwidth]{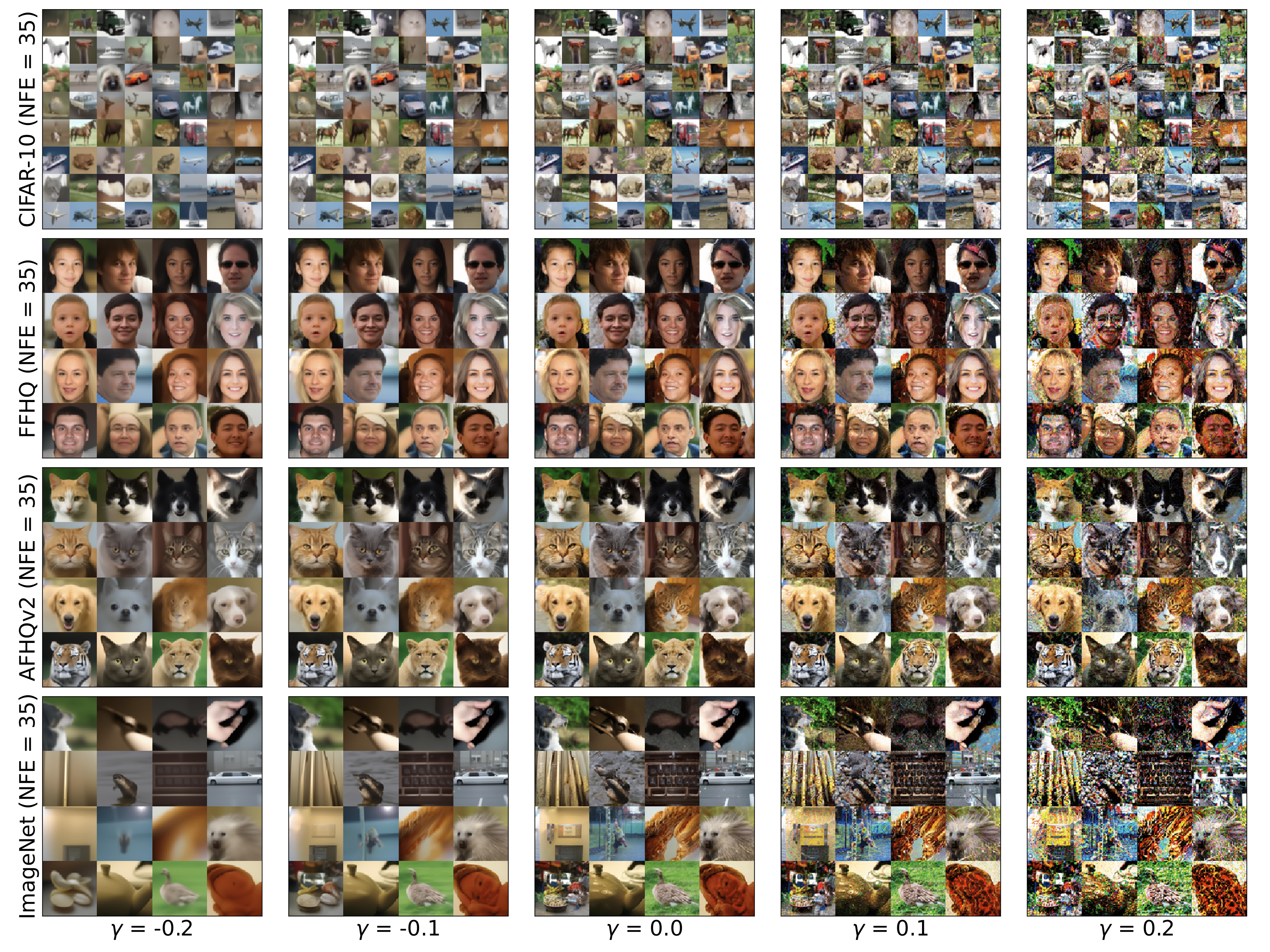}
    \caption{Generated images with different values of $\gamma$.}
    \label{fig:Varying_gamma}
\end{figure*}

\end{document}